\documentclass[10pt,twocolumn,letterpaper]{article}

\usepackage{cvpr}              

\newcommand{\TODO}[1]{\textbf{\color{red}[TODO: #1]}}
\renewcommand{\TODO}[1]{}

\usepackage{microtype}
\usepackage[accsupp]{axessibility}  

\renewcommand{\paragraph}[1]{\vspace{.5em}\noindent\textbf{#1.}}

\usepackage{soul}
\setuldepth{foobar}

\definecolor{cvprblue}{rgb}{0.21,0.49,0.74}
\usepackage[pagebackref,breaklinks,colorlinks,allcolors=cvprblue]{hyperref}

\definecolor{gold}{rgb}{0.8, 0.5, 0.2}
\definecolor{silver}{rgb}{0.66, 0.66, 0.66}

\usepackage{lipsum}
\usepackage{multirow, booktabs}
\usepackage[table,xcdraw]{xcolor}
\usepackage{color}
\usepackage{tcolorbox}
\tcbuselibrary{breakable}

\def\paperID{*****} 
\def\confName{CVPR}
\def\confYear{2026}

\title{\textsc{FPBench}: A Comprehensive Benchmark of Multimodal Large Language Models for Fingerprint Analysis}

\author{Ekta Gavas\\
New York University\\
{\tt\small eg4131@nyu.edu}
\and
Sudipta Banerjee\\
University of Wyoming\\
{\tt\small sbanerj3@uwyo.edu}
\and
Chinmay Hegde\\
New York University\\
{\tt\small chinmay.h@nyu.edu}
\and
Nasir Memon\\
New York University\\
{\tt\small memon@nyu.edu}
}

\begin{document}
\maketitle


\begin{abstract}
Multimodal LLMs (MLLMs) are capable of performing complex data analysis, visual question answering, generation, and reasoning tasks. However, their ability to analyze biometric data is relatively underexplored. In this work, we investigate the effectiveness of MLLMs in understanding fine structural and textural details present in fingerprint images. To this end, we design a comprehensive benchmark, \textsc{FPBench}, to evaluate 20 MLLMs (open-source and proprietary models) across 7 real and synthetic datasets on a suite of 8 biometric and forensic tasks (e.g., pattern analysis, fingerprint verification, real versus synthetic classification, etc.) using zero-shot and chain-of-thought prompting strategies. We further fine-tune vision and language encoders on a subset of open-source MLLMs to demonstrate domain adaptation. \textsc{FPBench} is a novel benchmark designed as a first step towards developing foundation models in fingerprints. Our findings indicate fine-tuning of vision and language encoders improves the performance by $\approx 7\%-39\%$. Our codes are available at \url{https://github.com/Ektagavas/FPBench}.

\end{abstract}    

\section{Introduction}
\label{sec:intro}
\begin{figure}[htb!]
        \centering
        \includegraphics[width=\columnwidth]{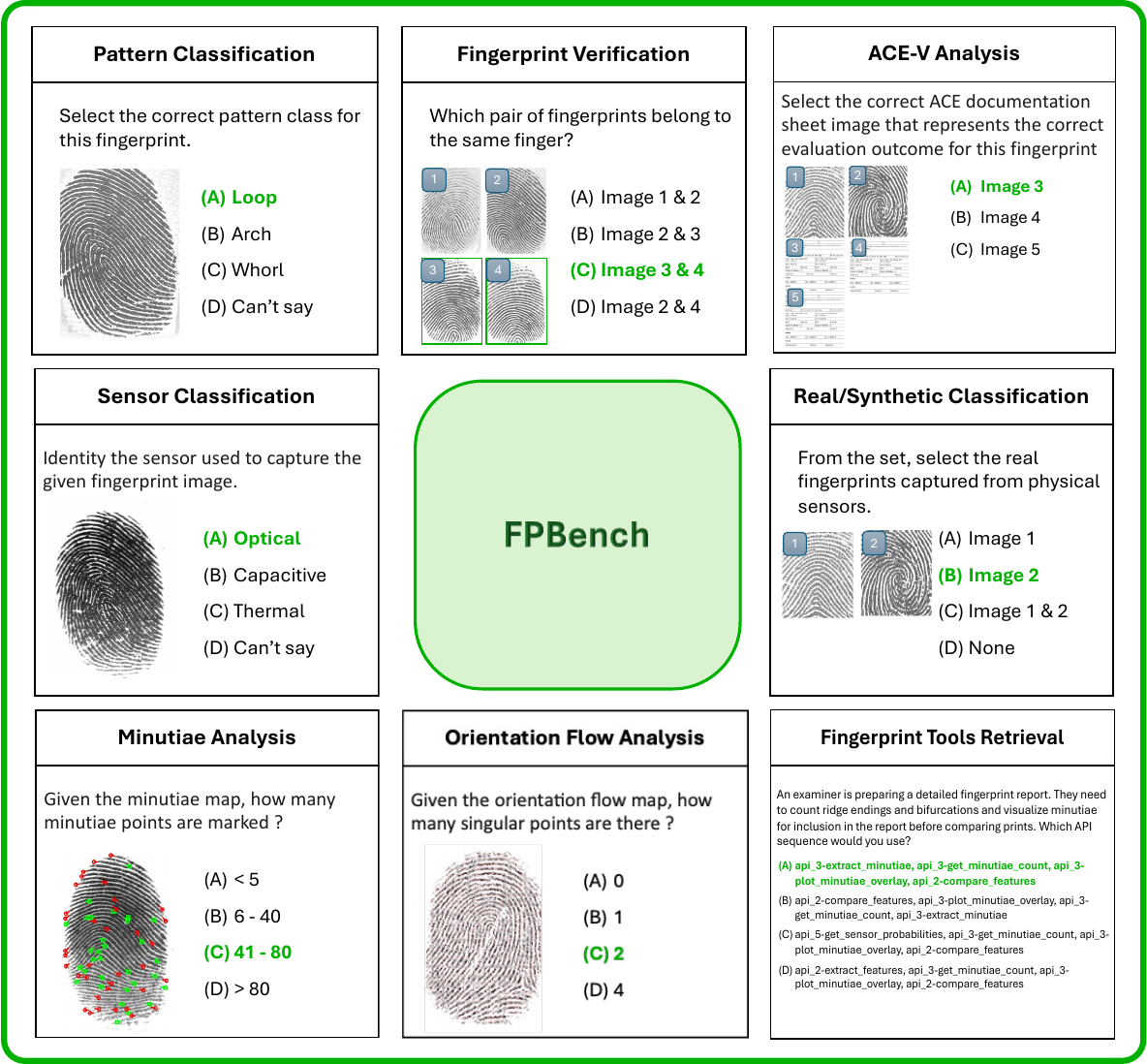}
        \captionof{figure}{\textsc{FPBench}: Overview of proposed benchmark for fingerprint analysis using MLLMs. We present examples of prompts curated for each task to evaluate the vision and language capabilities of MLLMs in fingerprint-based biometric and forensic tasks.}
        \label{fig:tasks}
\end{figure}

Multimodal LLMs (MLLMs) are inherently capable of processing multiple modalities (\textit{e.g.}, image, audio, text, tabular data, genomic sequences, etc.), to produce captions, descriptions, and chain-of-thought (CoT) guided logical reasoning with impressive details \cite{yin2024survey, wang2024comprehensive, sarto2025image}. Benchmarks are therefore important in systematically evaluating their performance, providing standardized tasks, datasets, and protocols for comparative evaluation. Existing benchmarks target general-purpose vision–language reasoning (e.g., VQAv2, GQA \cite{hudson2019gqa}), scientific diagrams (e.g., SciBench \cite{wang2023scibench}), charts (e.g., ChartQA \cite{masry2022chartqa}), documents (e.g.,  \cite{mathew2021docvqa, tito2021document}), and medical imaging (e.g., MedBench \cite{liu2024medbench}). Despite these advances, MLLMs are relatively underexplored in biometric modalities, particularly \textit{fingerprints}.
In this paper, we focus on this specialized domain that is significantly more complex than conventional tasks and has a strong potential to be a transformative technology; see Fig~\ref{fig:tasks} and Tab~\ref{fig:tasks}. 


\textbf{Motivation.} Fingerprints represent one of the earliest and most widely deployed biometric modalities with critical applications in forensics, law enforcement, border security, and personal authentication~\cite{Introtobiom}. Unlike general visual understanding tasks, fingerprint analysis presents unique domain-specific challenges, such as non-linear deformation, noisy and partial impressions, sensor variations, which are further compounded by presentation attacks~\cite{NIST_DigitalID}. Customized algorithms exist to address these challenges, which often require separate task-specific models. We intuit that the current generation of vision-language models that are trained on billions of data points may be capable of complementing the specialized task pipelines in fingerprint analysis. Thus, we propose \textsc{FPBench} to gain important insights into the \textit{effectiveness of MLLMs as assistive reasoning systems} and demonstrate the feasibility of \textit{domain adaptation with fine-tuning}.

\begin{table}
        \centering
        \captionof{table}{Five categories and eight tasks examined by \textsc{FPBench}.}
        \label{tab:tasks}
        \resizebox{0.85\columnwidth}{!}{%
        \begin{tabular}{l|l}
\hline
\multicolumn{1}{c}{\textbf{Categories}} & \multicolumn{1}{c}{\textbf{Tasks}} \\ \hline \hline
\multirow{2}{*}{Feature Analysis}       & Orientation Flow Analysis \\
 & Minutiae Analysis  \\ \hline
\multirow{2}{*}{Recognition}            & Pattern Classification \\
& Fingerprint Verification \\ \hline
\multirow{2}{*}{\begin{tabular}[c]{@{}l@{}}Source Attribution \&\\ Integrity Validation\end{tabular}} & Sensor Classification \\   & Real/Synthetic Classification   \\ \hline
Forensic Analysis & ACE-V Analysis  \\ \hline
Tool Use  & Fingerprint Tools Retrieval   \\ \hline                       
\end{tabular}
}
\end{table}

\indent\textbf{Objective:} \textit{Our goal is to assess whether MLLMs can effectively analyze fingerprints for specialized tasks.} So, we have designed a standardized benchmark, \textsc{FPBench}, to evaluate MLLMs for fingerprints, advancing the use of foundation models in biometrics. 
To the best of our knowledge, \textsc{FPBench} is a novel benchmark that focuses on specialized fingerprint analysis tasks using reasoning models.\\
\indent\textbf{Approach.} We examine the performance of a total of 20 MLLMS on fingerprint analytic tasks with \textit{zero-shot} and \textit{chain-of-thought (CoT)} prompting. 
We divide the eight fingerprint tasks considered in this benchmark into five categories. For each task, we prepared prompts based on multiple-choice questions (MCQ) using fingerprint datasets. 
Our \textbf{main contributions} are as follows:
\begin{enumerate}
    \item We introduce \textsc{FPBench}, a \textit{first-of-its-kind }evaluation benchmark for fingerprint understanding using MLLMs. We evaluate 18 open-source and 2 proprietary MLLM models on a suite of biometric and forensic tasks using real and synthetic datasets.
    \item We investigate \textit{zero-shot} and \textit{chain-of-thought (CoT)} prompt design strategies to guide the MLLMs to perform complex feature analysis for verification and classification tasks using structured reasoning.  
    \item We further investigate the impact of \textit{fine-tuning of vision and language encoders} for successful domain adaptation and compare our findings with state-of-the-art baselines in fingerprint analysis.
\end{enumerate}



\section{Related Work}


\subsection{Fingerprint Analysis}
Recent progress in fingerprint recognition relies on the use of deep learning, \textit{e.g.}, CNNs in \cite{tang2017fingernet, engelsma2019learning, nguyen2018robust, takahashi2020fingerprint, darlow2017fingerprint, gavas821enhancement} and transformer-based networks \cite{tandon2022transformer} for feature extraction and matching. FingerNet \cite{tang2017fingernet} combines ridge orientation and minutiae extraction in a unified CNN framework, while DeepPrint \cite{engelsma2019learning} learns compact, fixed-length representations for fast, scalable verification. MinutiaeNet \cite{nguyen2018robust} performs automated minutiae map extraction. Transformer-based approaches such as \cite{tandon2022transformer} further combine global and local representations with a smart matching process. Despite these advances, most deep learning approaches treat fingerprint recognition as a purely visual task, lacking the ability to provide natural language justifications or structured reasoning needed in forensic workflows, such as ACE-V analysis performed by NIST SWGFAST~\cite{SWGFAst_Doc9,SWGFAst_Doc10}.
To date, the lack of explainability restricts the possibility of an AI forensic agent capable of making automated decisions. Our work addresses this gap by evaluating whether MLLMs can assist in fingerprint tasks while providing coherent reasoning. 


\subsection{MLLM Benchmarks}
Recent advances in large language models have enabled multi-modal reasoning across images, video, audio, and text, supporting tasks like image/video captioning, visual question answering, image/video classification, object detection, text recognition, etc. Models such as InternVL \cite{chen2024far, chen2024internvl,zhu2025internvl3}, Gemma \cite{team2024gemma, team2024gemma2, gemma_2025}, LLaVA \cite{liu2023visual,liu2024improved} and Qwen-VL \cite{bai2023qwenvlversatilevisionlanguagemodel} have demonstrated strong performance on general vision-language benchmarks like MMBench \cite{liu2024mmbench}, MMStar\cite{chen2024we}, MathVista \cite{lu2023mathvista}, MMMU \cite{yue2024mmmu}. These benchmarks evaluate models across various aspects like captioning, knowledge-based reasoning, and scene understanding. There is a growing interest in adapting MLLMs to scientific and medical domains, requiring high interpretability and precision. Benchmarks like ScienceQA \cite{lu2022learn}, MathVista \cite{lu2023mathvista}, and PMC-VQA \cite{zhang2023pmc} in the biomedical domain show the adaptability of MLLMs towards reasoning in domain-specific tasks. 

\subsection{Applications of MLLM in Biometric Tasks}
While multimodal LLMs (MLLMs) have traditionally been evaluated on generic vision-language tasks, their application to specialized fields like biometrics is limited. \cite{sony2025benchmarking} examined the use of embeddings from foundational visual encoders for diverse biometric tasks in face and iris images. In face images, SHIELD \cite{shi2025shield} was an early attempt to benchmark MLLMs for face spoofing and forgery detection. EMO-LLaMA~\cite{xing2024emo} enhanced facial expression understanding capabilities. Face-MLLM \cite{sun2024face} handled multiple specialized face perception tasks. \cite{narayan2025facexbench} designed a comprehensive benchmark of 28 models, called FaceXBench, to evaluate the capabilities of MLLMs on 14 facial analytic tasks. FaceLLM \cite{shahreza2025facellm} demonstrated LLM capabilities on face-centric tasks, and introduced question-answer pairs dataset called FairFaceGPT. \cite{farmanifard2024chatgpt} explored the ChatGPT-4 capabilities for iris recognition, suggesting potential use of LLMs in specialized biometric applications. MLLMs have been used for heterogeneous face recognition~\cite{MLLM_2026_hetero}, automated emotional regulation~\cite{MLLM_2025_emotion} and person re-identification~\cite{MLLM_2025_reid}; refer to the survey of foundation models in biometrics~\cite{FoundationBiom_2025}. However, to the best of our knowledge, no study exists till date that evaluates MLLMs on fingerprint images. \textbf{Therefore, we present \textsc{FPBench}, the first open-source benchmark of MLLMs for biometric and forensic fingerprint analysis.}

\section{\textsc{FPBench}}
\subsection{Overview}
\textsc{FPBench} is a benchmark designed to systematically evaluate the ability of MLLMs to interpret fingerprint images. It spans eight tasks that collectively reflect key stages of fingerprint analysis, ranging from fundamental pattern recognition to high-level domain understanding (ACE-V). Inspired by FaceXBench \cite{narayan2025facexbench}, we formulate each task as a set of MCQ-style questions, ensuring consistent evaluation across diverse models. The benchmark explicitly tests visual understanding, spatial reasoning, and understanding of intricate geometric and textural cues from ridge patterns, minutiae, orientation, and sensor characteristics. 
\textsc{FPBench} provides a unified framework for both quantitative and qualitative assessment of MLLMs in fingerprint understanding.

\subsection{Categories and Tasks}
We selected a total of eight tasks across five categories that are evaluated in \textsc{FPBench}. We discuss them as follows:
\begin{enumerate}

    \item \textbf{Feature-Level Analysis:} 
    \begin{enumerate}

        \item \textbf{Orientation Flow Analysis:} This task evaluates understanding of ridge flow dynamics to
        \textit{interpret singularities} (cores/deltas). Each query probes
        the model to identify the number of singular points or fingerprint pattern (Level 1 features), by providing orientation map~\cite{hong1998fingerprint} overlaid on fingerprint image. Correct reasoning requires spatial and geometric understanding, like curvature and ridge direction.

    \begin{tcolorbox}[colback=lightgray!20, colframe=gray!50, rounded corners,breakable,before skip=8pt, after skip=8pt]

\small \textit{\textcolor{teal}{ Sample prompts for orientation flow analysis:}}\\
\textbf{User Q1:} In the given fingerprint orientation flow image, how many singular points can you find?
\\ 
\textbf{Input Image(s):} Image 1\\
(A) 2  (B) 0  (C) 1  (D) 3 \\
\textbf{User Q2:} Identify the pattern type of the given fingerprint orientation flow image.
\\ 
\textbf{Input Image(s):} Image 1\\
(A) Whorl (B) Can't Say (C) Arch (D) Loop
\end{tcolorbox}

        \item \textbf{Minutiae Analysis:} This task assesses the ability of MLLMs in counting \textit{the number of minutiae} (ridge endings and ridge bifurcations). Each question presents either fingerprint image(s) or minutiae maps (obtained using MINDTCT \cite{ko2007user}) overlaid on fingerprint images and prompts the model to compute the number of minutiae expressed within a range of values (e.g., 0-15, 16–35, 36–50, 51-70, $>$70). This task assesses how well MLLMs understand fine-grained local ridge structures (Level 2 features), thus complementing global reasoning task of orientation flow analysis.
        
        \begin{tcolorbox}[colback=lightgray!20, colframe=gray!50, rounded corners,breakable,before skip=8pt, after skip=8pt]

\small \textit{\textcolor{teal}{Sample prompts for minutiae analysis:}}\\
\textbf{User Q1:} From the minutiae map shown, choose the range that represents total count of minutiae. 
\\ 
\textbf{Input Image(s):} Image 1\\
(A) 51-70 
(B) 0-15 
(C) 16-35 
(D) more than 70 \\
\textbf{User Q2:} Choose the image that has ridge bifurcations in range 51-70? 
\\ 
\textbf{Input Image(s):} Image 1, 2, and 3\\
(A) Image 1 and Image 3 
(B) Image 3 
(C) Image 2 
(D) All of them 
\end{tcolorbox}
\end{enumerate}
    \item \textbf{Recognition:} 
        \begin{enumerate}[leftmargin=*]
            \item \textbf{Pattern Classification:} This task evaluates the model’s ability to build upon orientation flow analysis for the advanced task of \textit{fingerprint pattern classification}.  This task is a fundamental step in forensic analysis (Level 1 details). Each question presents 
            fingerprint image(s) that queries the model to classify them as ``loop", ``whorl", ``arch", or none of these, emphasizing global visual understanding.

\begin{tcolorbox}[colback=lightgray!20, colframe=gray!50, rounded corners, breakable, before skip=8pt, after skip=8pt
]

\small \textit{\textcolor{teal}{Sample prompts for pattern classification:}}\\
\textbf{User Q1:} Choose the correct fingerprint pattern category for this image.\\
\textbf{Input Image(s):} Image 1
\\ 
(A) Loop
(B) Whorl
(C) Arch
(D) Can't say\\
\textbf{User Q2:} Choose the image that corresponds to the loop fingerprint pattern. 
\\ 
\textbf{Input Image(s):} Image 1, 2 and 3\\
(A) Image 1 and Image 3 
(B) Image 3 
(C) Image 2 and Image 3 
(D) None of them


\end{tcolorbox}
            \item \textbf{Fingerprint Verification:} This task assesses a model’s capability to \textit{compare two fingerprint impressions and decide match} (same finger) \textit{or a non-match} (different finger). Each question contains multiple images, and the model needs to distinguish between the matching and non-matching pairs reflecting the verification phase in authentication and forensic comparison. Balanced genuine and impostor pairs with both \textit{easy} and \textit{difficult} examples are considered (\textit{e.g.}, cross-sensor, partial or noisy impressions). It measures the model’s visual reasoning ability to detect correspondence between ridge flows (global) and minutiae arrangement (local). 
        
\begin{tcolorbox}[colback=lightgray!20, colframe=gray!50, rounded corners,before skip=8pt, after skip=8pt]
\small \textit{\textcolor{teal}{Sample prompt for fingerprint verification:}}\\
\textbf{User Q1:} Which two fingerprints were likely captured from the same finger? 
\\ 
\textbf{Input Image(s):} Image 1, 2, 3 and 4\\
(A) Image 3 and Image 4\\
(B) Image 1 and Image 3\\
(C) Image 1 and Image 2\\
(D) Image 1 and Image 4
\end{tcolorbox}
\end{enumerate}

\item \textbf{Source Attribution \& Integrity Validation:}
    \begin{enumerate}
        \item \textbf{Sensor Classification:} Sensor classification evaluates the ability of MLLMs to \textit{identify fingerprint acquisition sensors} (\textit{e.g.}, optical, capacitive, thermal). Each question presents fingerprint image(s) and queries about the acquisition device. The task probes the model’s sensitivity to sensor-specific characteristics embedded in the image in the form of texture and contrast cues for forensic source attribution.
        \begin{tcolorbox}[colback=lightgray!20, colframe=gray!50, rounded corners,breakable, before skip=8pt, after skip=8pt]

\small \textit{\textcolor{teal}{Sample prompts for sensor classification:}}\\
\textbf{User Q1:} What is the sensor type that captured this fingerprint? \\ 
\textbf{Input Image(s):} Image 1\\
(A) Optical 
(B) Can't Say 
(C) Thermal 
(D) Capacitive 
\\
\textbf{User Q2:} Count how many fingerprint images are captured using thermal sensor. \\ 
\textbf{Input Image(s):} Image 1, 2, and 3\\
(A) 2 
(B) 0 
(C) 1 
(D) 3 
\end{tcolorbox}
        \item \textbf{Real or Synthetic Fingerprint Classification:} This task assesses whether models can \textit{distinguish between bona fide/real and synthetic fingerprints} which can be potentially used for spoof detection. Each question prompts the model to classify the image(s) as ``Real" or ``Synthetic". The models try to discern between generative model-based artifacts and sensor-specific cues for classification. 

\begin{tcolorbox}[colback=lightgray!20, colframe=gray!50, rounded corners,before skip=8pt,breakable, after skip=8pt]
\small \textit{\textcolor{teal}{Sample prompt for real vs synth classification:}}\\
\textbf{User Q1:} Select the fingerprints that are not real.
\\ 
\textbf{Input Image(s):} Image 1, 2, 3, and 4\\
(A) Image 2 and Image 3\\
(B) Image 2, 3, and 4 \\
(C) None of them \\
(D) All of them 
\end{tcolorbox}
    

\end{enumerate}

\item \textbf{Forensic Analysis:} 
Here, we consider ACE-V analysis.\\
    \textbf{ACE-V Analysis:} The goal of forensic fingerprint examination, termed as ACE-V (Analysis, Comparison, Evaluation and Verification), is to determine/exclude the identity of a print \cite{SWGFAst_Doc9, SWGFAst_Doc10}. In the \textit{Analysis} phase, Level 1 (friction ridge flow, pattern, singularity points), Level 2 (minutiae) and Level 3 features (pores, incipient ridges, creases, scars, etc.) are analysed. \textit{Comparison} phase involves side-by-side comparison to determine feature agreement. In the \textit{Evaluation} phase, examiners conclude individualization (match), exclusion (non-match) or inconclusiveness. The decision is validated by another examiner in the \textit{Verification} phase. The ACE-V Analysis task in \textsc{FPBench} probes the model’s ability to emulate the structured reasoning used for \textit{deriving the outcome of `Evaluation' step in the ACE-V methodology}. Questions present ACE-V sheet images with a pair of fingerprints and ask the model to choose the most likely outcome according to ACE-V guidelines. For example, if two prints show sufficient agreement in Level-1 and Level-2 details, then the outcome is \textit{Individualization}, while disagreement between features yields an \textit{Exclusion} outcome and finally, insufficient analysis produces an \textit{Inconclusive} outcome. 
\begin{tcolorbox}[colback=lightgray!20, colframe=gray!50, rounded corners,breakable,before skip=8pt, after skip=8pt]

\small \textit{\textcolor{teal}{Sample prompt for ACE-V analysis:}}\\
\textbf{User Q1:} Based on the fingerprint images given as Image 1 and Image 2, analyze them carefully and select the correct ACE worksheet image from images 3, 4 and 5 that accurately represents the correct evaluation outcome for this fingerprint pair.
\\ 
\textbf{Input Image(s):} Image 1, 2, 3, 4, 5\\
(A) Image 3 
(B) Image 4 
(C) Image 5 \\ 

\textit{where Image 1 and 2 are fingerprints and Images 3, 4, 5 are ACE sheets with conclusion as `Individualization', `Exclusion' and `Inconclusive'.}
\end{tcolorbox}

\item \textbf{Tool Use:} 
Here, we consider retrieving fingerprint tools.\\
    \textbf{Fingerprint Tools Retrieval:} This task examines the model’s capability for \textit{determining the correct sequence of steps/tools for a specific task}. We present the model with a biometric or forensic task and ask the model to select the option corresponding to the correct order, as well as the least number of API calls needed to complete a given task. This assesses the future potential of MLLMs to be deployed as an end-to-end AI agent for automated fingerprint analysis.

\begin{tcolorbox}[colback=lightgray!20, colframe=gray!50, rounded corners,breakable,before skip=8pt, after skip=8pt]

\small \textit{\textcolor{teal}{Sample prompt for tool retrieval:}}\\
\textbf{User Q1:} In a biometric research lab, a system must determine if a fingerprint sample is synthetically generated before extracting its features for further evaluation. The sample must be enhanced first. Pick the correct API sequence with only necessary APIs. 
\\
(A) api\_4-get\_orientation\_angles, api\_6-get\_real\_ probabilities, api\_2-extract\_features \\
(B) api\_8-enhance\_image, api\_6-predict\_real\_ synthetic, api\_2-extract\_features \\
(C) api\_2-extract\_features, api\_6-get\_real\_ probabilities, api\_8-enhance\_image \\
(D) api\_8-enhance\_image, api\_1-get\_pattern\_ probabilities, api\_2-extract\_features 
\end{tcolorbox}
\end{enumerate}





\section{Experiments}
\subsection{Datasets}
To comprehensively evaluate \textsc{FPBench}, we curated data from several publicly available datasets. The datasets ensure variations in terms of fingerprint quality, acquisition conditions, and task relevance. 
\begin{itemize}
    \item \textit{FVC2000}~\cite{maio2002fvc2000},  \textit{FVC2002}~\cite{maio2002fvc2002} and \textit{FVC2004}~\cite{maio2004fvc2004}: Introduced in Fingerprint Verification Competitions (FVC), each dataset contains 440 fingers with 8 impressions each, captured across various sensors/conditions and employed for orientation flow analysis, minutiae analysis, pattern classification and verification.
    \item \textit{NIST SD302d} ~\cite{fiumara2019nist}: NIST SD302 has plain, rolled and touch-free impressions captured from various devices, making it challenging with partial and noisy impressions. We use the subset 302d containing 5141 fingerprint images acquired from four different auxiliary devices.
    \item \textit{NIST SD301a} ~\cite{fiumara2018nist}: NIST SD301 spans diverse image quality and acquisition conditions, captured by fingerprint experts. The subset 301a contains a total of 4366 images.
    \item \textit{GenPrint} ~\cite{grosz2024universal}: This large-scale synthesized dataset contains 150K highly realistic, diverse fingerprint impressions with acquisition, sensor, fingerprint class, and quality variations. It is particularly valuable for synthetic classification and cross-domain generalization studies. 
    \item \textit{Anguli}~\cite{anguli}: This dataset contains 10K fingerprints, generated using an open-source handcrafted fingerprint generator `Anguli', also used for real vs. synthetic classification and analyzes robustness across generative models.
\end{itemize}  


\subsection{Models}
We used 2 proprietary models: GPT-5 \cite{gpt5} and Gemini 2.5 Pro \cite{comanici2025gemini} API versions to ensure privacy \cite{geminiterms, openaiterms}. We used 18 open-source models. The choice of models is based on state-of-the-art performance on other biometric modalities~\cite{narayan2025facexbench}. We divide open-source models into three major categories based on parameter size: 

\begin{itemize}
    \item \textbf{Open-Source MLLMs (\textless4B parameters)}: LLaVA-OneVision-0.5b-OV\cite{li2024llava}, Qwen3-VL-2b-Instruct\cite{Qwen3-VL};
    \item \textbf{Open-Source MLLMs (4B-13B parameters)}: Gemma3-4b\cite{gemma_2025}, Chameleon-7b\cite{team2024chameleon}, LLaVA-v1.5-7b\cite{liu2023visual}, LLaVA-NeXT-Interleave-7b\cite{li2024llavanext}, LLaVA-OneVision-7b-SI\cite{li2024llava}, LLaVA-OneVision-7b-OV\cite{li2024llava}, DeepSeek-VL-7b\cite{lu2024deepseekvl}, Qwen3-VL-8b-Instruct\cite{Qwen3-VL}, Monkey-Chat\cite{li2024monkey},
    Idefics2-8b\cite{laurenccon2024matters}, InternVL3-8b\cite{zhu2025internvl3}, Idefics-9b-Instruct\cite{laurenccon2023obelics}, and Gemma3-12b\cite{gemma_2025};
    \item \textbf{Open-Source MLLMs (\textgreater13B parameters)}: LLaVA-v1.5-13b\cite{liu2023visual}, Qwen3-VL-13b-Instruct\cite{Qwen3-VL} and InternVL3-38b\cite{zhu2025internvl3}.
\end{itemize}
 
\textbf{SOTA benchmark evaluation.} We compare the findings of \textsc{FPBench} with existing fingerprint-specific  SOTA models. We used NIST-BOZORTH3 \cite{ko2007user} as a minutiae-based matcher and a transformed-based deep-learning matcher \cite{tandon2022transformer} for verification. We used Gemini 2.5 Pro as the SOTA model for remaining tasks since there was no single best-performing algorithm that could serve as the benchmark.
\begin{table*}
\centering
\caption{Accuracy (\%) of models in \textsc{FPBench} with \textcolor{blue}{zero-shot} / \textcolor{blue}{CoT} prompting. We categorize the open-source models in three categories based on parameter size: (a) \textless{}4B parameters, (b) 4B-13B parameters, (c) \textgreater{}13B parameters, and (d) proprietary models. The highest and second-highest performance values in each category/setting are highlighted in \textcolor{gold}{gold} and \textcolor{silver}{silver}, respectively. Values in parentheses indicate the number of questions in each task.}
\label{tab:result}
\resizebox{0.95\textwidth}{!}{%
\begin{tabular}{|llllllllll|}
\hline
\rowcolor[HTML]{FFCCC9}
\multicolumn{1}{|l|}{} &
  \multicolumn{1}{c|}{\textbf{\begin{tabular}[c]{@{}l@{}}Overall \\ (5000)\end{tabular}}} &
  \multicolumn{1}{c|}{\textbf{\begin{tabular}[c]{@{}c@{}}Pattern \\ (800)\end{tabular}}} &
  \multicolumn{1}{c|}{\textbf{\begin{tabular}[c]{@{}c@{}}Minutiae \\ (800)\end{tabular}}} &
  \multicolumn{1}{c|}{\textbf{\begin{tabular}[c]{@{}c@{}}Orientation \\ (400)\end{tabular}}} &
  \multicolumn{1}{c|}{\textbf{\begin{tabular}[c]{@{}c@{}}Verification \\ (800)\end{tabular}}} &
  \multicolumn{1}{c|}{\textbf{\begin{tabular}[c]{@{}c@{}}Sensor \\ (800)\end{tabular}}} &
  \multicolumn{1}{c|}{\textbf{\begin{tabular}[c]{@{}c@{}}Real/Synthetic \\ (600)\end{tabular}}} &
  \multicolumn{1}{c|}{\textbf{\begin{tabular}[c]{@{}c@{}}ACE-V Analysis \\ (600)\end{tabular}}} &
  \multicolumn{1}{c|}{\textbf{\begin{tabular}[c]{@{}c@{}}Tools Retrieval \\ (100)\end{tabular}}} \\ \hline
\multicolumn{1}{|l|}{\textbf{Random Choice}} &
  \multicolumn{1}{c|}{25.40} &
  \multicolumn{1}{c|}{26.23} &
  \multicolumn{1}{c|}{25.99} &
  \multicolumn{1}{c|}{24.38} &
  \multicolumn{1}{c|}{25.00} &
  \multicolumn{1}{c|}{27.98} &
  \multicolumn{1}{c|}{25.67} &
  \multicolumn{1}{c|}{25.91} &
  \multicolumn{1}{c|}{22.00} \\
\multicolumn{1}{|l|}{\textbf{Frequent Choice}} &
  \multicolumn{1}{c|}{28.90} &
  \multicolumn{1}{c|}{27.58} &
  \multicolumn{1}{c|}{28.08} &
  \multicolumn{1}{c|}{29.06} &
  \multicolumn{1}{c|}{29.23} &
  \multicolumn{1}{c|}{27.23} &
  \multicolumn{1}{c|}{27.33} &
  \multicolumn{1}{c|}{34.72} &
   \multicolumn{1}{c|}{28.00}\\ \hline
\multicolumn{1}{|l|}{\textbf{SOTA Models}} &
  \multicolumn{1}{c|}{\begin{tabular}{@{}cc@{}}
      57.88 \\ \hline
      61.44 \\ 
    \end{tabular}
    } &
  \multicolumn{1}{c|}{46.92} &
  \multicolumn{1}{c|}{48.77} &
  \multicolumn{1}{c|}{45.57} &
  \multicolumn{1}{c|}{
    \begin{tabular}{@{}c|c@{}}
      Bozorth3\cite{ko2007user} & 61.32 \\ \hline
      Deep-learning-based\cite{tandon2022transformer} & 89.80 \\ 
    \end{tabular}%
} &
  \multicolumn{1}{c|}{49.50} &
  \multicolumn{1}{c|}{35.26} &
  \multicolumn{1}{c|}{79.73} &
   \multicolumn{1}{c|}{96.00}\\ \hline
\multicolumn{10}{|c|}{
\cellcolor[HTML]{E5F7FC}
\textbf{Open source MLLMs (\textless 4B parameters)}} \\ \hline
\multicolumn{1}{|l|}{\textbf{LLaVA-OneVision-0.5b-OV\cite{li2024llava}}} &
\multicolumn{1}{c|}{34.05 / 31.68} & 
\multicolumn{1}{c|}{34.73 / 31.40} & 
\multicolumn{1}{c|}{23.28 / 26.72} & 
\multicolumn{1}{c|}{33.50 / 30.30} & 
\multicolumn{1}{c|}{29.23 / 28.23} & 
\multicolumn{1}{c|}{44.65 / 38.68} & 
\multicolumn{1}{c|}{26.32 / 24.56} & 
\multicolumn{1}{c|}{34.72 / 33.55} & 
\multicolumn{1}{c|}{46.00 / 40.00} \\ 
\multicolumn{1}{|l|}{\textbf{Qwen3-VL-2b-Instruct\cite{Qwen3-VL}}} &
\multicolumn{1}{c|}{34.38 / 34.78} & 
\multicolumn{1}{c|}{35.84 / 31.28} & 
\multicolumn{1}{c|}{27.09 / 28.94} & 
\multicolumn{1}{c|}{34.98 / 37.19} & 
\multicolumn{1}{c|}{32.96 / 34.70} & 
\multicolumn{1}{c|}{20.52 / 29.85} & 
\multicolumn{1}{c|}{27.72 / 24.56} & 
\multicolumn{1}{c|}{29.90 / 34.22} & 
\multicolumn{1}{c|}{66.00 / 57.50}  \\ \hline
\multicolumn{10}{|c|}{
\cellcolor[HTML]{E5F7FC}
\textbf{Open source MLLMs (4B - 13B parameters)}} \\ \hline
\multicolumn{1}{|l|}{\textbf{Gemma3-4b\cite{gemma_2025}}} &
\multicolumn{1}{c|}{37.50 / 33.34} & 
\multicolumn{1}{c|}{41.87 / 29.80} & 
\multicolumn{1}{c|}{28.20 / 23.52} & 
\multicolumn{1}{c|}{38.42 / 34.48} & 
\multicolumn{1}{c|}{29.73 / 35.82} & 
\multicolumn{1}{c|}{41.42 / 26.00} & 
\multicolumn{1}{c|}{22.81 / 20.18} & 
\multicolumn{1}{c|}{33.55 / 35.88} & 
\multicolumn{1}{c|}{64.00 / 61.00} \\
\multicolumn{1}{|l|}{\textbf{Chameleon-7b\cite{team2024chameleon}}} &
\multicolumn{1}{c|}{23.77 / 21.89} & 
\multicolumn{1}{c|}{24.75 / 25.74} & 
\multicolumn{1}{c|}{22.54 / 22.66} & 
\multicolumn{1}{c|}{26.85 / 24.14} & 
\multicolumn{1}{c|}{19.90 / 16.92} & 
\multicolumn{1}{c|}{23.01 / 21.77} & 
\multicolumn{1}{c|}{20.53 / 11.93} & 
\multicolumn{1}{c|}{18.60 / 15.95} & 
\multicolumn{1}{c|}{34.00 / 36.00} \\
\multicolumn{1}{|l|}{\textbf{LLaVA-v1.5-7b\cite{liu2023visual}}} &
\multicolumn{1}{c|}{31.23 / 27.24} & 
\multicolumn{1}{c|}{36.70 / 23.40} & 
\multicolumn{1}{c|}{24.63 / 24.38} & 
\multicolumn{1}{c|}{28.82 / 20.20} & 
\multicolumn{1}{c|}{24.00 / 20.27} & 
\multicolumn{1}{c|}{40.42 / 36.44} & 
\multicolumn{1}{c|}{17.37 / 16.14} & 
\multicolumn{1}{c|}{33.89 / 34.05} & 
\multicolumn{1}{c|}{44.00 / 43.00} \\
\multicolumn{1}{|l|}{\textbf{LLaVA-NeXT-Interleave-7b\cite{li2024llavanext}}} &
\multicolumn{1}{c|}{37.98 / 38.14} & 
\multicolumn{1}{c|}{30.79 / 29.56} & 
\multicolumn{1}{c|}{25.12 / 24.01} & 
\multicolumn{1}{c|}{30.54 / 27.34} & 
\multicolumn{1}{c|}{38.31 / 38.43} & 
\multicolumn{1}{c|}{41.92 / 41.54} & 
\multicolumn{1}{c|}{28.42 / 26.67} & 
\multicolumn{1}{c|}{33.72 / 33.55} & 
\multicolumn{1}{c|}{75.00 / 84.00} \\
\multicolumn{1}{|l|}{\textbf{LLaVA-OneVision-7b-SI\cite{li2024llava}}} &
\multicolumn{1}{c|}{42.00 / 37.12} & 
\multicolumn{1}{c|}{\textcolor{silver}{48.89} / 37.32} & 
\multicolumn{1}{c|}{29.56 / 24.01} & 
\multicolumn{1}{c|}{40.15 / 34.48} & 
\multicolumn{1}{c|}{32.46 / 28.98} & 
\multicolumn{1}{c|}{34.33 / 33.21} & 
\multicolumn{1}{c|}{30.35 / 27.72} & 
\multicolumn{1}{c|}{32.23 / 31.23} & 
\multicolumn{1}{c|}{88.00 / 80.00} \\
\multicolumn{1}{|l|}{\textbf{LLaVA-OneVision-7b-OV\cite{li2024llava}}} &
\multicolumn{1}{c|}{39.32 / 37.62} & 
\multicolumn{1}{c|}{36.58 / 33.62} & 
\multicolumn{1}{c|}{29.56 / 34.73} & 
\multicolumn{1}{c|}{39.16 / 36.21} & 
\multicolumn{1}{c|}{30.35 / 31.59} & 
\multicolumn{1}{c|}{28.36 / 29.35} & 
\multicolumn{1}{c|}{30.00 / 27.37} & 
\multicolumn{1}{c|}{29.57 / 29.07} & 
\multicolumn{1}{c|}{91.00 / 79.00} \\
\multicolumn{1}{|l|}{\textbf{DeepSeek-VL-7b\cite{lu2024deepseekvl}}} &
\multicolumn{1}{c|}{38.32 / 36.94} & 
\multicolumn{1}{c|}{46.31 / \textcolor{gold}{48.28}} & 
\multicolumn{1}{c|}{33.99 / \textcolor{silver}{35.22}} & 
\multicolumn{1}{c|}{32.51 / 29.31} & 
\multicolumn{1}{c|}{26.62 / 26.37} & 
\multicolumn{1}{c|}{49.50 / 49.38} & 
\multicolumn{1}{c|}{31.75 / 31.40} & 
\multicolumn{1}{c|}{33.89 / 33.55} & 
\multicolumn{1}{c|}{52.00 / 42.00} \\
\multicolumn{1}{|l|}{\textbf{Qwen3-VL-8b-Instruct\cite{Qwen3-VL}}} &
\multicolumn{1}{c|}{49.42 / 46.68} & 
\multicolumn{1}{c|}{\textcolor{gold}{50.49} /  \textcolor{silver}{42.86}} & 
\multicolumn{1}{c|}{35.71 / 28.69} & 
\multicolumn{1}{c|}{42.36 / 40.39} & 
\multicolumn{1}{c|}{47.89 / 41.79} & 
\multicolumn{1}{c|}{\textcolor{gold}{58.08} / \textcolor{gold}{53.73}} & 
\multicolumn{1}{c|}{38.07 / 29.30} & 
\multicolumn{1}{c|}{31.73 / 45.68} & 
\multicolumn{1}{c|}{91.00 / 91.00}  \\
\multicolumn{1}{|l|}{\textbf{Monkey-Chat\cite{li2024monkey}}} &
\multicolumn{1}{c|}{30.99 / 30.60} & 
\multicolumn{1}{c|}{34.36 / 33.25} & 
\multicolumn{1}{c|}{22.41 / 21.31} & 
\multicolumn{1}{c|}{32.27 / 30.30} & 
\multicolumn{1}{c|}{30.10 / 28.73} & 
\multicolumn{1}{c|}{34.20 / 32.96} & 
\multicolumn{1}{c|}{18.07 / 17.19} & 
\multicolumn{1}{c|}{36.54 / 37.04} & 
\multicolumn{1}{c|}{40.00 / 44.00}  \\
\multicolumn{1}{|l|}{\textbf{Idefics2-8b\cite{laurenccon2024matters}}} &
\multicolumn{1}{c|}{34.31 / 33.68} & 
\multicolumn{1}{c|}{39.41 / 33.50} & 
\multicolumn{1}{c|}{24.88 / 27.34} & 
\multicolumn{1}{c|}{28.08 / 28.82} & 
\multicolumn{1}{c|}{25.37 / 24.63} & 
\multicolumn{1}{c|}{36.69 / 37.31} & 
\multicolumn{1}{c|}{22.81 / 22.46} & 
\multicolumn{1}{c|}{31.23 / 31.40} & 
\multicolumn{1}{c|}{66.00 / 64.00} \\
\multicolumn{1}{|l|}{\textbf{InternVL3-8b\cite{zhu2025internvl3}}} &
\multicolumn{1}{c|}{46.55 / 38.78} & 
\multicolumn{1}{c|}{41.26 / 35.71} & 
\multicolumn{1}{c|}{29.93 / 26.23} & 
\multicolumn{1}{c|}{37.19 / 36.21} & 
\multicolumn{1}{c|}{44.78 / 37.94} & 
\multicolumn{1}{c|}{48.01 / 33.21} & 
\multicolumn{1}{c|}{34.74 / 21.58} & 
\multicolumn{1}{c|}{46.51 / 37.38} & 
\multicolumn{1}{c|}{90.00 / 82.00} \\
\multicolumn{1}{|l|}{\textbf{Idefics-9b-Instruct\cite{laurenccon2023obelics}}} &
\multicolumn{1}{c|}{30.94 / 29.79} & 
\multicolumn{1}{c|}{33.13 / 29.68} & 
\multicolumn{1}{c|}{23.40 / 22.66} & 
\multicolumn{1}{c|}{30.54 / 29.56} & 
\multicolumn{1}{c|}{20.65 / 23.76} & 
\multicolumn{1}{c|}{43.53 / 40.55} & 
\multicolumn{1}{c|}{21.05 / 21.23} & 
\multicolumn{1}{c|}{34.22 / 33.89} & 
\multicolumn{1}{c|}{41.00 / 37.00} \\
\multicolumn{1}{|l|}{\textbf{Gemma3-12b\cite{gemma_2025}}} &
\multicolumn{1}{c|}{47.03 / 40.06} & 
\multicolumn{1}{c|}{48.15 / 36.82} & 
\multicolumn{1}{c|}{31.53 / 27.22} & 
\multicolumn{1}{c|}{43.10 / 37.44} & 
\multicolumn{1}{c|}{44.78 / 40.30} & 
\multicolumn{1}{c|}{47.89 / 36.32} & 
\multicolumn{1}{c|}{31.93 / 26.49} & 
\multicolumn{1}{c|}{44.85 / 42.86} & 
\multicolumn{1}{c|}{84.00 / 73.00} \\ \hline
\multicolumn{10}{|c|}{
\cellcolor[HTML]{E5F7FC}
\textbf{Open source MLLMs (\textgreater 13B parameters)}} \\ \hline
\multicolumn{1}{|l|}{\textbf{LLaVA-v1.5-13b\cite{liu2023visual}}} &
\multicolumn{1}{c|}{31.14 / 26.88} & 
\multicolumn{1}{c|}{37.32 / 27.71} & 
\multicolumn{1}{c|}{24.88 / 26.97} & 
\multicolumn{1}{c|}{27.83 / 20.20} & 
\multicolumn{1}{c|}{24.00 / 22.76} & 
\multicolumn{1}{c|}{31.47 / 25.37} & 
\multicolumn{1}{c|}{17.37 / 18.77} & 
\multicolumn{1}{c|}{34.22 / 34.22} & 
\multicolumn{1}{c|}{52.00 / 39.00} \\
\multicolumn{1}{|l|}
  {\textbf{Qwen3-VL-32b-Instruct\cite{Qwen3-VL}}} &
\multicolumn{1}{c|}{52.39 / \textcolor{silver}{47.23}} & 
\multicolumn{1}{c|}{47.66 / 32.64} & 
\multicolumn{1}{c|}{36.82 / 30.67} & 
\multicolumn{1}{c|}{45.07 / 35.71} & 
\multicolumn{1}{c|}{52.61 / 50.00} & 
\multicolumn{1}{c|}{52.86 / \textcolor{silver}{46.14}} & 
\multicolumn{1}{c|}{32.11 / 30.88} & 
\multicolumn{1}{c|}{\textcolor{silver}{58.97} / \textcolor{silver}{58.80}} & 
\multicolumn{1}{c|}{93.00 / 93.00}\\
\multicolumn{1}{|l|}
  {\textbf{InternVL3-38b\cite{zhu2025internvl3}}} &
\multicolumn{1}{c|}{\textcolor{silver}{52.86} / 46.24} & 
\multicolumn{1}{c|}{42.86 / 38.67} & 
\multicolumn{1}{c|}{31.65 / 31.03} & 
\multicolumn{1}{c|}{\textcolor{silver}{45.81} / \textcolor{silver}{41.87}} & 
\multicolumn{1}{c|}{\textcolor{gold}{56.84} / \textcolor{silver}{50.37}} & 
\multicolumn{1}{c|}{\textcolor{silver}{57.59} / 36.69} & 
\multicolumn{1}{c|}{\textcolor{silver}{39.82} / 32.28} & 
\multicolumn{1}{c|}{51.33 / 47.01} & 
\multicolumn{1}{c|}{\textcolor{gold}{97.00} / 92.00} \\
  \hline
\multicolumn{10}{|c|}{
\cellcolor[HTML]{FFFFC7}
\textbf{Proprietary MLLMs}} \\ \hline
\multicolumn{1}{|l|}{\textbf{GPT-5\cite{gpt5}}} &
\multicolumn{1}{c|}{50.20 / \textcolor{gold}{54.45}} & 
\multicolumn{1}{c|}{46.18 / 41.87} & 
\multicolumn{1}{c|}{\textcolor{silver}{38.79} / \textcolor{gold}{45.94}} & 
\multicolumn{1}{c|}{\textcolor{gold}{50.49} / \textcolor{gold}{43.35}} & 
\multicolumn{1}{c|}{44.90 / \textcolor{gold}{54.73}} & 
\multicolumn{1}{c|}{49.50 / 43.03} & 
\multicolumn{1}{c|}{\textcolor{gold}{46.49} / \textcolor{gold}{51.23}} & 
\multicolumn{1}{c|}{29.24 / \textcolor{gold}{59.47}} & 
\multicolumn{1}{c|}{\textcolor{silver}{96.00} / \textcolor{gold}{96.00}} \\
\multicolumn{1}{|l|}{\textbf{Gemini 2.5 Pro\cite{comanici2025gemini}}} &
\multicolumn{1}{c|}{\textcolor{gold}{57.01} / 44.42} & 
\multicolumn{1}{c|}{46.92 / 34.73} & 
\multicolumn{1}{c|}{\textcolor{gold}{48.77} / 29.31} & 
\multicolumn{1}{c|}{45.57 / 34.73} & 
\multicolumn{1}{c|}{\textcolor{silver}{54.35} / 36.82} & 
\multicolumn{1}{c|}{49.50 / 39.68} & 
\multicolumn{1}{c|}{35.26 / \textcolor{silver}{40.88}} & 
\multicolumn{1}{c|}{\textcolor{gold}{79.73} / 44.19} & 
\multicolumn{1}{c|}{\textcolor{silver}{96.00} / \textcolor{silver}{95.00}} \\ \hline
\multicolumn{1}{|l|}{\textbf{AVERAGE}} &
  \multicolumn{1}{c|}{40.01 / 36.88} &
\multicolumn{1}{c|}{40.21 / 33.89} & 
\multicolumn{1}{c|}{29.63 / 28.08} & 
\multicolumn{1}{c|}{36.67 / 32.61} & 
\multicolumn{1}{c|}{35.49 / 33.66} & 
\multicolumn{1}{c|}{41.67 / 36.56} & 
\multicolumn{1}{c|}{28.65 / 26.14} & 
\multicolumn{1}{c|}{37.93 / 37.65} & 
\multicolumn{1}{c|}{70.30 / 66.42}\\ \hline
\end{tabular}%
}
\end{table*}
\begin{table*}
\centering
\caption{Results of fine-tuning vision and language layers in 3 open-source models with varying number of parameters across 7 tasks in \textsc{FPBench}. `Before FT' results indicate results prior to fine-tuning (derived from Tab \ref{tab:result}) and `After FT' denotes \textcolor{blue}{zero-shot} prompting results after fine-tuning. We observe consistent improvement in performance across the board, indicating successful domain adaptation.}
\label{tab:ft-results}
\resizebox{1.00\textwidth}{!}{%
\begin{tabular}{|l|ll|ll|ll|ll|ll|ll|ll|}
\hline
\rowcolor[HTML]{FFCCC9}
\multicolumn{1}{|c|}{} & \multicolumn{2}{c|}{\textbf{Pattern}} & \multicolumn{2}{c|}{\textbf{Minutiae}} & \multicolumn{2}{c|}{\textbf{Orientation}} & \multicolumn{2}{c|}{\textbf{Verification}} & \multicolumn{2}{c|}{\textbf{Sensor}} & \multicolumn{2}{c|}{\textbf{Real/Synthetic}} & \multicolumn{2}{c|}{\textbf{ACE-V Analysis}} \\ \cline{2-15} 
\multicolumn{1}{|c|}{\multirow{-2}{*}
{\cellcolor[HTML]{FFCCC9}\textbf{Model/Task}}} & \multicolumn{1}{l|}{\cellcolor[HTML]{D3D3D3}Before FT} & After FT & \multicolumn{1}{l|}{\cellcolor[HTML]{D3D3D3}Before FT} & After FT & \multicolumn{1}{l|}{\cellcolor[HTML]{D3D3D3}Before FT} & After FT & \multicolumn{1}{l|}{\cellcolor[HTML]{D3D3D3}Before FT} & After FT & \multicolumn{1}{l|}{\cellcolor[HTML]{D3D3D3}Before FT} & After FT & \multicolumn{1}{l|}{\cellcolor[HTML]{D3D3D3}Before FT} & After FT & \multicolumn{1}{l|}{\cellcolor[HTML]{D3D3D3}Before FT} & After FT \\ \hline
\textbf{\cellcolor[HTML]{E5F7FC}Qwen3-VL-8b-Instruct} & 
\multicolumn{1}{l|}{\cellcolor[HTML]{D3D3D3}50.49} &
  58.25 &
  \multicolumn{1}{l|}{\cellcolor[HTML]{D3D3D3}35.71} &
  52.96 &
  \multicolumn{1}{l|}{\cellcolor[HTML]{D3D3D3}42.36} &
  53.20 &
  \multicolumn{1}{l|}{\cellcolor[HTML]{D3D3D3}47.89} &
  62.69 &
  \multicolumn{1}{l|}{\cellcolor[HTML]{D3D3D3}58.08} &
  64.43 &
  \multicolumn{1}{l|}{\cellcolor[HTML]{D3D3D3}38.07} &
  54.50 &
  \multicolumn{1}{l|}{\cellcolor[HTML]{D3D3D3}31.73} &
  46.84 \\ \hline
\textbf{\cellcolor[HTML]{E5F7FC}Gemma3-12b} &
  \multicolumn{1}{l|}{\cellcolor[HTML]{D3D3D3}48.15} &
  47.79 &
  \multicolumn{1}{l|}{\cellcolor[HTML]{D3D3D3}31.53} &
  43.72 &
  \multicolumn{1}{l|}{\cellcolor[HTML]{D3D3D3}43.10} &
  55.17 &
  \multicolumn{1}{l|}{\cellcolor[HTML]{D3D3D3}44.78} &
  53.61 &
  \multicolumn{1}{l|}{\cellcolor[HTML]{D3D3D3}47.89} &
  51.12 &
  \multicolumn{1}{l|}{\cellcolor[HTML]{D3D3D3}31.93} &
  38.17 &
  \multicolumn{1}{l|}{\cellcolor[HTML]{D3D3D3}44.85} &
  56.15 \\ \hline
\textbf{\cellcolor[HTML]{E5F7FC}Qwen3-VL-32b-Instruct} &
  \multicolumn{1}{l|}{\cellcolor[HTML]{D3D3D3}47.66} &
  61.21 &
  \multicolumn{1}{l|}{\cellcolor[HTML]{D3D3D3}36.82} &
  59.36 &
  \multicolumn{1}{l|}{\cellcolor[HTML]{D3D3D3}45.07} &
  58.37 &
  \multicolumn{1}{l|}{\cellcolor[HTML]{D3D3D3}52.61} &
  68.53 &
  \multicolumn{1}{l|}{\cellcolor[HTML]{D3D3D3}52.86} &
  74.13 &
  \multicolumn{1}{l|}{\cellcolor[HTML]{D3D3D3}32.11} &
  71.00 &
  \multicolumn{1}{l|}{\cellcolor[HTML]{D3D3D3}58.97} &
  78.74 \\ \hline
\end{tabular}%
}
\end{table*}
\subsection{Prompting Strategies for Evaluation}
We evaluate the models under two settings: (a) zero-shot and (b) chain-of-thought (CoT)
prompting. We used the system prompt as: \textit{``You are an expert fingerprint examiner"}. In the \textit{zero-shot} setting, we provided images and the query as: \textit{``Please answer the question and provide only the correct option letter, e.g., A, B, C, D."}. In the \textit{CoT} setting, we guide the model to provide step-by-step reasoning for its response. So, we prepend the prompt with \textit{``Please think step by step and provide your reasoning before giving the final answer. Include the final correct answer option at the end of your answer"}. Following~\cite{narayan2025facexbench}, we parse the model prediction for zero-shot and detailed reasoning in CoT setting.  

\subsection{Implementation Details}
\textbf{Evaluation.} We first created questions using GPT and then manually filtered them to ensure variations while being relevant. 
We randomized the option order and ensured that each question has only one correct option. The number of questions in each task is indicated in parentheses in Tab \ref{tab:result}. We used a pretrained CNN to assign ground-truth pattern labels for pattern classification task wherever applicable. For tools retrieval task, we defined the API list
and generated questions using GPT-5 and Gemini 2.5 Pro and refined them based on correctness.\\ 
\textbf{Fine-tuning.} We selected three models with varying parameter size \textit{viz.} \textit{Qwen3-VL-8b}, \textit{Gemma3-12b} and \textit{Qwen3-VL-32b}. We fine-tuned the vision projector and LLM backbone using LORA~\cite{hu2022lora}. We created the training data ($\sim 1800-2100$ MCQ for each task) for fine-tuning which is distinct from the evaluation set. \\
\textbf{Resources}. We used a single H200 GPU for both fine-tuning and evaluation. The evaluation benchmark code is based on VLMEvalKit \cite{duan2024vlmevalkit}.
Training time : $\sim$10 mins for Qwen3-VL-8b, $\sim$18 mins for Gemma3-12b and $\sim$15 mins for Qwen3-VL-32b. Inference time: average $\sim$6s per sample (\textit{e.g.}, in Qwen3-VL-8b).


\section{Results}
\begin{figure}
    \centering
    \includegraphics[width=0.95\columnwidth]{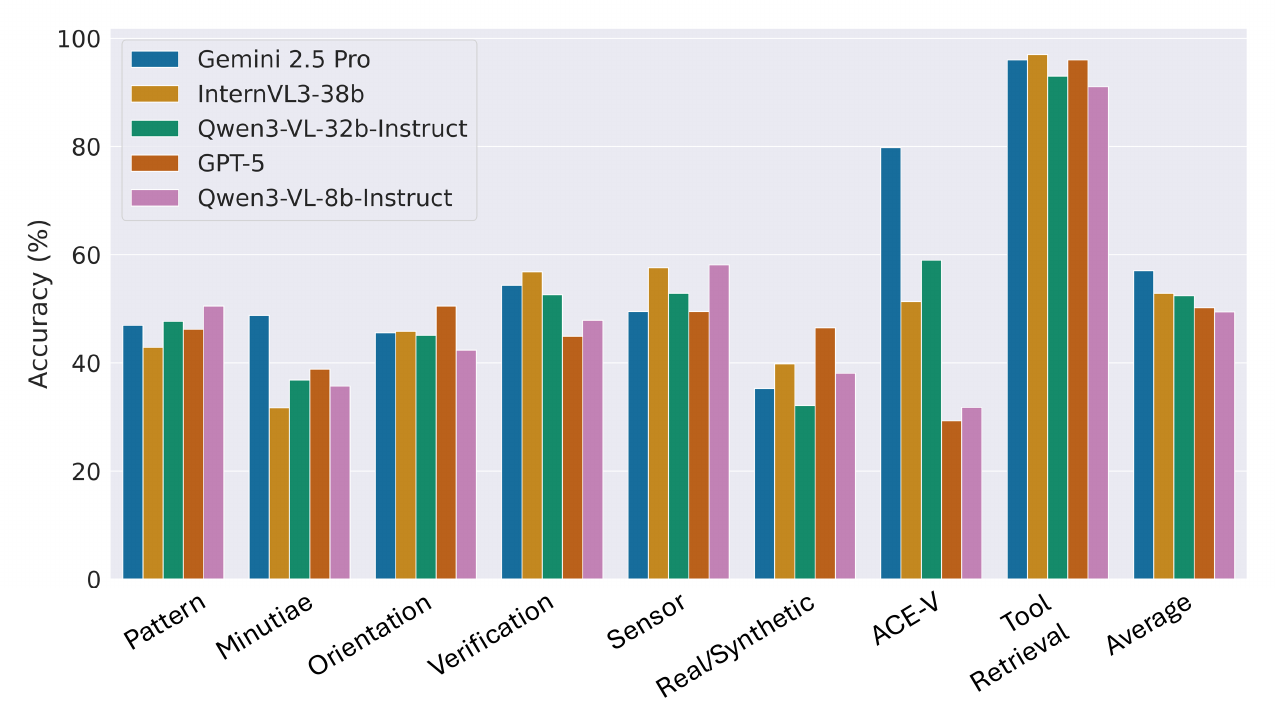}
    \caption{Accuracy (\%) of top-5 best performing models along with mean performance across all tasks}
    \label{fig:result_plotsA}
\end{figure}

We present the results of \textsc{FPBench} in terms of accuracy (\%) in Tab \ref{tab:result} for zero-shot and CoT prompting. \textit{Random Choice} accuracy is computed via selecting an option randomly from given options, while \textit{Frequent Choice} accuracy is computed by selecting the most frequently occurring correct option letter in that task. These baselines average 26–28\%, reflecting the difficulty of the MCQ setup. Nearly all MLLMs exceed these baselines, achieving above 50\% accuracy across most tasks ($2\times$ higher than random chance of 25\%), without \textit{any} domain-specific fine-tuning. Though only a few perform consistently well across all fingerprint tasks.  The performance reaches up to 96\% on tools retrieval task. Results indicate that although MLLMs currently lack in some aspects for specialized fingerprint analysis, it shows strong potential in advancing the state-of-the-art in the future. The zero-shot performance varies across models and parameter size. 
We further selected the top-5 best performing models and investigated their performance in Fig \ref{fig:result_plotsA}, 
indicating that the top-5 models perform marginally similar across most tasks except on the ACE-V analysis task where closed-source model Gemini 2.5 Pro outperforms the rest. Overall, Gemini 2.5 Pro leads the benchmark with 57.01\% accuracy, outperforming GPT-5 (50.20\%), InternVL3-38b (52.86\%), Qwen3-VL-32b (52.39\%) and Qwen3-VL-8b  (49.42\%) by a small margin but with stronger consistency across specialized tasks. 

\begin{figure}
    \centering
    \includegraphics[width=0.95\columnwidth]{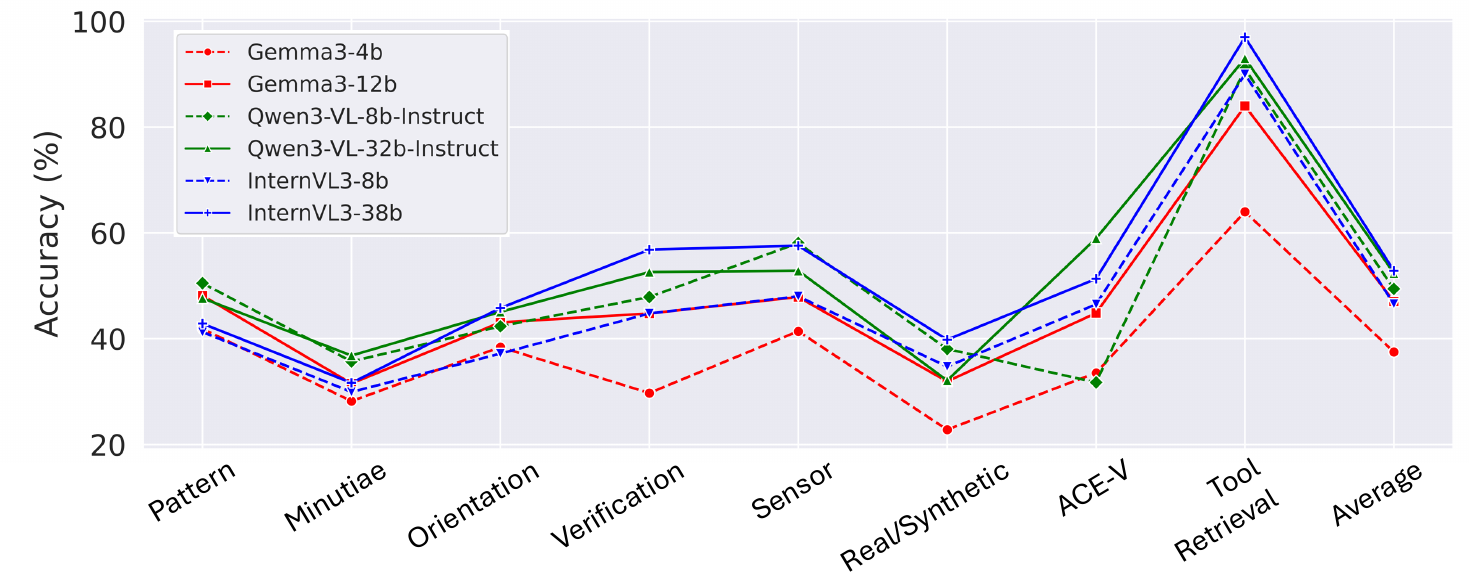}
    \caption{Performance (\%) variation of select models with change in model size (\#params). Solid lines represent larger model variants, while their lighter counterparts are indicated by dashed lines.}
    \label{fig:result_plotsB}
\end{figure}

\subsection{Evaluation performance across different tasks}
\label{subsec:tasks}

\quad \textbf{Orientation flow analysis.} In Tab \ref{tab:result}, we observe low accuracy (36.67\%) in orientation flow analysis. This task requires examining global ridge flow (pattern) and singular points (core/delta), which suggests that orientation flow overlays may provide limited complementary information for ridge flow understanding. In this task, GPT-5 (50.49\%) and InternVL3-38b (45.81\%) result in the highest and second-highest performance, respectively. 

\textbf{Minutiae analysis.} Minutiae analysis resulted in the second lowest performance (29.63\%) among eight tasks considered in this work. 
Gemini 2.5 Pro (48.77\%) emerges as the highest performing model, followed by GPT-5 (38.79\%). Both outperform open-source models, which achieve 22-36\% accuracy. This task involves understanding finer details in fingerprints, like identifying ridge endings and bifurcations. The results confirm the need for a domain-specific vision encoder and possibly image denoising and enhancement processing for fine-grained understanding. 

\textbf{Pattern classification.} We observe moderate performance (40.21\%) in pattern classification task. MLLMs are capable of extracting global ridge-flow features using generic vision encoders. In Tab \ref{tab:result}, we observe that over ten models cross the 40\% accuracy mark in this task, with Qwen3-VL-8b (50.49\%) leading them, showing high-level structure and texture understanding. 

\textbf{Fingerprint verification.} We observe average performance (35.49\%) in verification task that involves comparative analysis of global (Level-1) and local (Level-2) details for decision. As verification requires comparative reasoning, InternVL3-38b (56.84\%) and Gemini 2.5 Pro (54.35\%) excel here, likely due to superior cross-vision reasoning alignment. Yet, there is a huge scope of improvement compared to SOTA model benchmarks, BOZORTH3 \cite{ko2007user} (61.32\%) and deep-learning-based verification \cite{tandon2022transformer} (89.80\%). 

\textbf{Sensor classification}. We observe reasonable performance (41.67\%) in sensor classification task, suggesting that models can detect artifacts stemming from different sensors (e.g., optical vs. capacitive vs. thermal sensors). In this task, open-source models like Qwen3-VL-8b (58.08\%) and InternVL3-38b (57.59\%) perform better than closed-source models (49.50\%). 

\textbf{Real/Synthetic fingerprint classification.} We observed the \textit{lowest} performance (28.65\%) in this task, performing marginally above the frequent choice baseline (27.33\%), making it the \textit{most challenging} task in \textsc{FPBench}. Synthetic fingerprint cues are hard to capture when synthetic datasets like GenPrint mimic real-world data well. Further, MLLM vision encoders may lack sufficient fingerprint knowledge due to the unavailability of open-source data. Here, GPT-5 (46.49\%) performs better than InternVL3-38b (39.82\%) and Gemini 2.5 Pro (35.26\%). 

\textbf{ACE-V Analysis.} From Tab \ref{tab:result}, ACE-V Analysis task achieves modest performance (37.93\%). This task is particularly challenging as it requires fingerprint matching, a clear understanding of ACE-V terminology, and correctly parsing the ACE-V sheet. This process is performed manually by fingerprint experts and hence requires reasoning. Gemini 2.5 Pro performs significantly well (79.73\%) compared to other models. In contrast, GPT-5 attains only 29.24\%, warranting further exploration. 

\textbf{Tools retrieval.} In Tab \ref{tab:result} and Fig \ref{fig:result_plotsA}, MLLMs achieve the \textit{highest} performance (70.3\%) on the tool retrieval task. InternVL3-38b performs the best, securing 97\%, followed by GPT-5 and Gemini 2.5 Pro with 96\%. This suggests that tools retrieval is relatively the simplest task in \textsc{FPBench}. Results reinforce that MLLMs are good at understanding the fingerprint analysis pipeline and determining the correct sequence of operations, thus supporting the feasibility of foundational models for fingerprints using tool chaining.

\subsection{Fine-tuning performance}
We wanted to investigate whether MLLMs can gain \textit{domain knowledge} that will improve their understanding and subsequently their performance. So, we performed task-specific fine-tuning (FT) of the vision and LLM layers of three open-source models using LORA on various fingerprint tasks. We selected the models based on their performance in Tab~\ref{tab:result}. We evaluated the performance on 7 tasks (omitted Tools Retrieval as it was already achieving $\approx 95\%$ without fine-tuning). Tab \ref{tab:ft-results} highlights the benefits of fine-tuning MLLMs for fingerprint tasks in zero-shot setting. Qwen3-VL-32b performs the \textit{best} across all tasks. In pattern classification, Qwen3 models show improvement, with Qwen3-VL-32b achieving the highest accuracy (61.21\%), while Gemma3-12b shows marginal decrease of 0.3\%. In minutiae analysis task, Gemma3-12b achieves the lowest performance (43.72\%) compared to Qwen3 models (59.36\%), surpassing its performance prior to FT (31.53\%). The trend is similar to most tasks except orientation analysis and ACE-V, where Qwen3-VL-8b shows the least improvement (53.2\% and 46.84\% respectively). Qwen3-VL models show significant improvement with Qwen3-VL-32b achieving  68.53\% (verification), 74.13\% (sensor classification), and 71\% (real/synthetic classification). It achieves 78.74\% accuracy in ACE-V analysis task, which is comparable to the very large model Gemini 2.5 Pro without fine-tuning. Note here that the Qwen3 models outperform the BOZORTH3 (61.32\%) benchmark in verification task. This indicates that MLLMs can be improved with domain-specific fine-tuning to perform well on complex biometric and forensic tasks.

\subsection{Additional Analysis}
\subsubsection{Effect of size on MLLM performance}
From Tab \ref{tab:result}, the small-tier ($<$4B) open-source models perform variably (30-46\%), while the mid-tier Qwen3-VL-8b (49.42\%), InternVL3-8b (46.55\%), and Gemma3-12b (47.03\%) models perform reasonably well across all categories. Larger models ($>$13B) \textit{viz}., InternVL3-38b (52.86\%) and Qwen3-VL-32b (52.39\%), achieving comparable results with proprietary models like GPT-5 and Gemini 2.5 Pro. We observe higher performance with the increase in number of parameters within the same series, like Gemma3, Qwen3 or InternVL3 (see Fig \ref{fig:result_plotsB}).

\subsubsection{Zero-shot vs. Chain-of-Thought (CoT)} 
We summarized the CoT evaluation results (without fine-tuning) in Tab \ref{tab:result}. Some models benefit from CoT on a few tasks, while we observe lower performance overall if CoT is used without fine-tuning. Models like LLaVa and DeepSeek benefit from CoT prompting in the minutiae analysis task, while Gemma3-4b, LLaVa, Idefics-9b show performance gain in verification task. Both proprietary (GPT-5) and open-source models (Qwen3-VL-2b) indicate potential gains with CoT prompting. These findings align with the results reported in FaceXBench~\cite{narayan2025facexbench} on face images. We will explore how fine-tuning improves CoT evaluation in the future.



\subsection{Discussion}




We reviewed the model predictions for hallucinations and biases. Some models exhibited strong answer-selection biases, even with a randomized option order. For example, Idefics-9b never chose option A, Monkey-Chat avoided option B (ACE-V) and option D (Tool Retrieval), and multiple LLaVA variants consistently selected option A in ACE-V. Additionally, Chameleon-7b failed to predict for nearly 20\% of the questions (except in Tool Retrieval), likely due to its limited 4096-token context window. In single-image pattern and sensor classification tasks, we observed class-level biases with open-source and proprietary models favoring `whorl' and `loop' class respectively. Whereas `arch' was consistently under-predicted. In sensor classification, `optical' sensor category was predominantly predicted, except in LLaVA-OneVision-7b and Qwen3-VL-2b models, which favored `capacitive' class.
Currently, we do not explore the vulnerabilities of MLLMs in the presence of prompt injection and adversarial attacks. 

\textbf{SOTA baselines vs MLLMs.} While fingerprint SOTA baselines achieve higher performance (Tab \ref{tab:result} verification task), they are typically task-specific and lack interpretability and unified reasoning capabilities, required in a forensic ACE-V setup. Our goal is not to replace them, but to see if MLLMs can act as forensic assistants that integrate knowledge and reasoning across various fingerprint stages.

\textbf{Choice of MCQ-style evaluation.} We formulated the questions in \textsc{FPBench} in MCQ format, as it enables diverse evaluation of MLLMs under both positive and negative scenarios (\textit{e.g.}, when ``None of them" is correct), providing a fair comparison across different models. It allows flexibility for the model to select `Can't say' option (\textit{e.g.}, sensor classification) thus ensuring reliability.

\textbf{Reproducibility.} We wanted to investigate whether the responses provided by the models contain statistical variations. So, we report the mean and standard deviation of accuracy computed across three runs for three tasks in \textsc{FPBench} in Tab \ref{tab:std}. We observe reasonably low values of standard deviation. For inference, we set the parameters \texttt{temperature=0} or \texttt{do\_sample=False} in the open-source models to enforce determinism. Additional analysis is provided in \textcolor{black}{Supplementary Materials}.
\begin{table}
\centering
\caption{Mean and standard deviation of accuracy (\%) reported across three inference runs on three tasks in \textsc{FPBench}.}
\label{tab:std}
\resizebox{0.9\columnwidth}{!}{%
\begin{tabular}{|l|r|r|r|}
\hline
\rowcolor[HTML]{FFCCC9}
 & \multicolumn{1}{l|}{\textbf{Pattern}} & \multicolumn{1}{l|}{\textbf{Minutiae}} & \multicolumn{1}{l|}{\textbf{Orientation}} \\ \hline
\textbf{Gemma3-4b}             & 41.13 $\pm$ 1.607 & 28.20 $\pm$ 0.245 & 36.86 $\pm$ 1.353 \\ \hline
\textbf{LLaVA-v1.5-7b}         & 36.94 $\pm$ 1.371 & 24.99 $\pm$ 1.212 & 28.98 $\pm$ 0.748 \\ \hline
\textbf{LLaVA-OneVision-7b-SI} & 46.92 $\pm$ 2.701 & 29.27 $\pm$ 1.683 & 39.74 $\pm$ 0.711 \\ \hline
\textbf{Qwen3-VL-8b-Instruct}  & 50.53 $\pm$ 0.075 & 36.33 $\pm$ 0.621 & 43.03 $\pm$ 0.571 \\ \hline
\end{tabular}%
}
\end{table}


\section{Summary}
In this work, we designed \textsc{FPBench}, a novel benchmark for evaluating the capability of MLLMs to analyze fingerprint data. First, we selected a set of 20 MLLMs (18 open source and 2 proprietary) and then curated multiple choice questions (MCQ)-based queries focusing on 8 specialized tasks of biometric analysis (\textit{e.g.}, pattern classification, minutiae analysis, verification, etc.) and forensic examination (\textit{e.g.}, ACE-V analysis done by forensic fingerprint experts, real vs. synthetic fingerprint classification, etc.). Secondly, we evaluated the models via zero-shot prompting and through guided chain-of-thought (CoT) prompting; we then compared their performance with existing baselines. We observed that proprietary models (Gemini 2.5 Pro and GPT-5) emerged as winners across several tasks, followed by open-source models (InternVL3-38b and Qwen3-VL). Certain tasks, like real vs. synthetic fingerprint classification were too challenging for the MLLMs. Finally, we leveraged domain adaptation by simultaneously fine-tuning both vision and language encoders of selected open-source models. We observed a significant improvement in performance between $\approx 7\%-39\%$. Future work will focus on enabling user interaction for reliability and leveraging tool chaining for developing an AI-assisted agent for fingerprint analysis. We will further consider constrained generation to improve trustworthiness~\cite{learntorefuse}. 
{
    \small
    \bibliographystyle{ieeenat_fullname}
    \bibliography{main}
}







%



\def\paperID{*****} 
\def\confName{CVPR}
\def\confYear{2026}



\maketitlesupplementary
\appendix
\section{Statistics of evaluation question prompts}
\label{apdx:stats}
The cumulative statistics for the questions in \textsc{FPBench} evaluation is provided in Tab \ref{tab:stats}. 

\begin{table}[h]
\centering
\caption{ Key statistics of questions in \textsc{FPBench}.}
\label{tab:stats}
\resizebox{0.85\columnwidth}{!}{%
\begin{tabular}{|l|l|}
\hline
\textbf{Statistic}               & \textbf{Number} \\ \hline
Total questions                  & 4940            \\
Total categories                 & 5               \\
Total tasks                      & 8               \\
Public datasets used             & 6               \\ \hline
Questions with multiple images   & 3220 (65\%)     \\
Questions with single image      & 1620 (33\%)     \\
Questions with only text         & 100 (2\%)       \\ \hline
Total images in all questions    & 13654           \\
Unique number of images          & 9034           \\
Unique question templates        & 419             \\ \hline
Maximum question length          & 1275             \\
Maximum option length            & 160             \\
Average question length          & 71.59           \\
Average option length            & 6.80           \\ \hline
Total options in each question   & 3 or 4               \\
Frequency of A as correct option & 1330 (26.92\%)  \\
Frequency of B as correct option & 1261 (25.52\%)  \\
Frequency of C as correct option & 1299 (26.29\%)  \\
Frequency of D as correct option & 1050 (21.69\%)  \\ \hline
\end{tabular}%
}
\end{table}
\section{Dataset statistics}
\label{apdx:dataset_stats}
Tab \ref{tab:dataset-stats} represents the fingerprint dataset statistics used in \textsc{FPBench}. Note that the datasets were cleaned to remove any latent prints, palm prints, digital (RGB) fingerprint photos or 3D prints prior to using them in \textsc{FPBench}.

\begin{table}[!ht]
\centering
\caption{Statistics of fingerprint datasets used for \textsc{FPBench}}
\label{tab:dataset-stats}
\resizebox{\columnwidth}{!}{%
\begin{tabular}{|c|c|c|c|}
\hline
\textbf{Dataset} &
  \textbf{\begin{tabular}[c]{@{}c@{}}No. of \\ Fingers\end{tabular}} &
  \textbf{\begin{tabular}[c]{@{}c@{}}No. of \\ Impressions\end{tabular}} &
  \textbf{\begin{tabular}[c]{@{}c@{}}Total \\ Images\end{tabular}} \\ \hline
FVC2000 \cite{maio2002fvc2000}     & 440    & 8   & 3520    \\
FVC2002 \cite{maio2002fvc2002}     & 440    & 8   & 3520    \\
FVC2004 \cite{maio2004fvc2004}     & 440    & 8   & 3520    \\
NIST SD302d \cite{fiumara2019nist} & 2000  & 1-3 & 5141   \\
NIST SD301a \cite{fiumara2018nist} & 240   &  1-15   & 4366   \\
GenPrint \cite{grosz2024universal}    & 10000 & 15  & 150000 \\
Anguli \cite{anguli}      & 10000 & -   & 10000  \\ \hline
\end{tabular}%
}
\end{table}

\section{Example ACE-V sheet}
\begin{figure*}[h]
    \centering
    \includegraphics[width=0.9\textwidth]{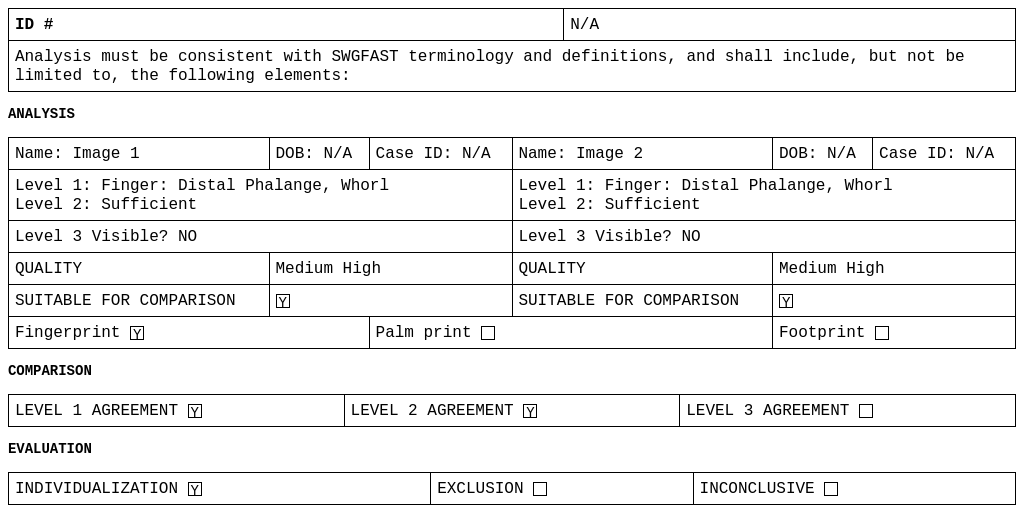}
    \caption{Example ACE-V sheet from the ACE-V analysis task on a sample fingerprint pair in \textsc{FPBench}. The template is referenced from Document 9 of SWGFAST documentations \cite{SWGFAst_Doc9}.}
    \label{fig:ace-sheet}
\end{figure*}

Fig \ref{fig:ace-sheet} shows an example ACE-V sheet for fingerprint analysis and evaluation for a pair of prints. The sheet concludes ``Individualization" as Level 1 and Level 2 features are in agreement.

\section{Proprietary models API config}
\label{apdx:apiconfig}
\begin{table}[]
\caption{Configuration settings for closed-source models in \textsc{FPBench}}
\label{tab:config}
\resizebox{\columnwidth}{!}{%
\begin{tabular}{|c|cc|}
\hline
\multirow{2}{*}{\textbf{Model}} & \multicolumn{2}{c|}{\textbf{zero-shot/chain-of-thought}}                             \\ \cline{2-3} 
 & \multicolumn{1}{c|}{\textbf{Reasoning}}                   & \textbf{maxOutputTokens} \\ \hline
GPT-5 \cite{gpt5}                           & \multicolumn{1}{c|}{reasoning\_effort=`minimal'/`medium'} & 32 / 128,000                  \\ \hline
Gemini 2.5 Pro \cite{comanici2025gemini}                   & \multicolumn{1}{c|}{thinkingBudget=128/dynamic}           & 136 / 65,536                 \\ \hline
\end{tabular}
}
\end{table}

Tab \ref{tab:config} refers to API configuration for proprietary models GPT-5 and Gemini 2.5 Pro in zero-shot and CoT settings.

\section{Additional Results}
\label{apdx:results}
\begin{figure*}[]
    \centering
    \includegraphics[width=0.8\textwidth]{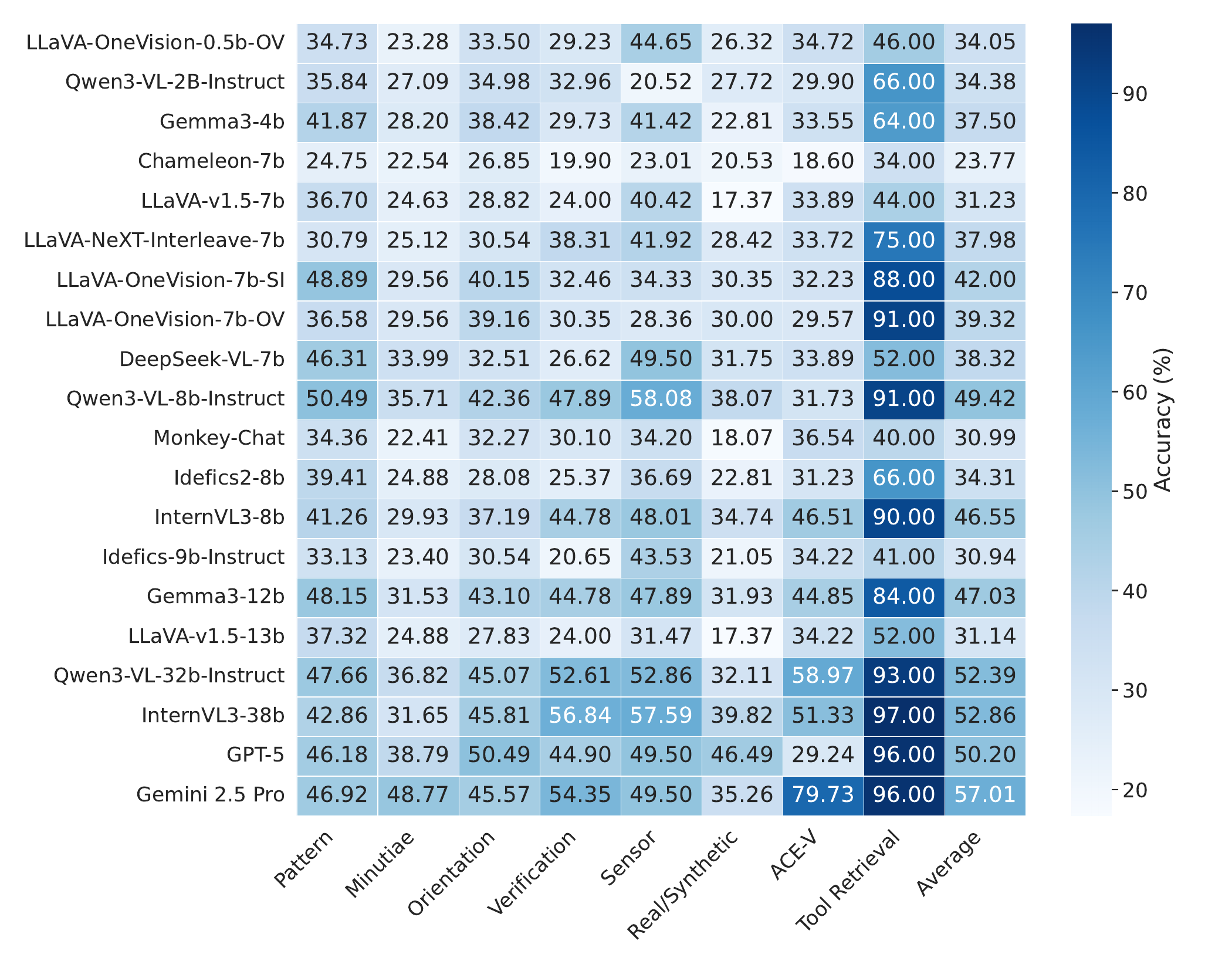}
    \caption{Accuracy (\%) of all models across various fingerprint tasks presented in the form of a heatmap. The Tool Retrieval task appears to be the best-performing task across a majority of the models, whereas all the models struggle to distinguish between real and synthetic fingerprints on the Real vs. Synthetic classification task.}
    \label{fig:plot_a}
\end{figure*}

\begin{figure*}
    \centering
    \includegraphics[width=0.60\textwidth]{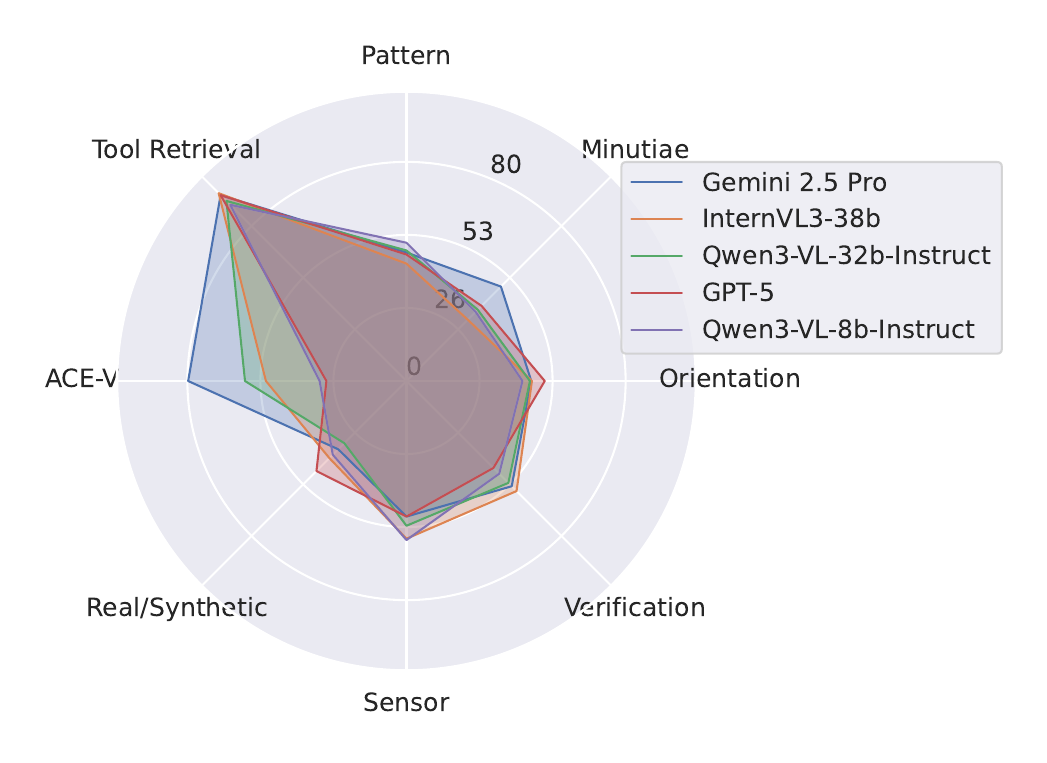}
    \caption{Performance (\%) of top-5 best performing models across all tasks on zero-shot prompting.}
    \label{fig:plot_radar}
\end{figure*}

The change in performance under different evaluation settings is shown in Tab \ref{tab:result-cot}. Fig \ref{fig:plot_a} depicts the performance of all the models across all the tasks in the form of a heatmap. This helps in understanding the individual and average performance of the models across all the tasks. The radar plot in Fig \ref{fig:plot_radar} indicates the accuracy with increasing value as one moves away from the center (0\%) towards the outer periphery (100\%); each axis on the concentric circle corresponds to a single task. This depicts that the model with a larger area (in all directions) has an overall better performance across all categories and tasks. 
\begin{table*}[t]
\centering
\caption{Change in performance (accuracy $\%$) under chain-of-thought compared to zero-shot evaluation setting in \textsc{FPBench}. Negative values suggest a decrease in performance in the chain-of-thought as compared to the zero-shot setting.}
\label{tab:result-cot}
\resizebox{0.95\textwidth}{!}{%
\begin{tabular}{|llllllllll|}
\hline
\rowcolor[HTML]{FFCCC9}
\multicolumn{1}{|l|}{} &
  \multicolumn{1}{c|}{\textbf{\begin{tabular}[c]{@{}l@{}}Overall \\ (5000)\end{tabular}}} &
  \multicolumn{1}{c|}{\textbf{\begin{tabular}[c]{@{}c@{}}Pattern \\ (800)\end{tabular}}} &
  \multicolumn{1}{c|}{\textbf{\begin{tabular}[c]{@{}c@{}}Minutiae \\ (800)\end{tabular}}} &
  \multicolumn{1}{c|}{\textbf{\begin{tabular}[c]{@{}c@{}}Orientation \\ (400)\end{tabular}}} &
  \multicolumn{1}{c|}{\textbf{\begin{tabular}[c]{@{}c@{}}Verification \\ (800)\end{tabular}}} &
  \multicolumn{1}{c|}{\textbf{\begin{tabular}[c]{@{}c@{}}Sensor \\ (800)\end{tabular}}} &
  \multicolumn{1}{c|}{\textbf{\begin{tabular}[c]{@{}c@{}}Real/Synthetic \\ (600)\end{tabular}}} &
  \multicolumn{1}{c|}{\textbf{\begin{tabular}[c]{@{}c@{}}ACE-V Analysis \\ (600)\end{tabular}}} &
  \multicolumn{1}{c|}{\textbf{\begin{tabular}[c]{@{}c@{}}Tools Retrieval \\ (100)\end{tabular}}} \\ \hline
\multicolumn{10}{|c|}{
\cellcolor[HTML]{E5F7FC}
\textbf{Open source MLLMs (\textless 4B parameters)}} \\ \hline
\multicolumn{1}{|l|}{\textbf{LLaVA-OneVision-0.5b-OV}} &
  \multicolumn{1}{c|}{-18.99} &
\multicolumn{1}{c|}{-3.33} & 
\multicolumn{1}{c|}{3.44} & 
\multicolumn{1}{c|}{-3.2} & 
\multicolumn{1}{c|}{-1.0} & 
\multicolumn{1}{c|}{-5.97} & 
\multicolumn{1}{c|}{-1.76} & 
\multicolumn{1}{c|}{-1.17} & 
\multicolumn{1}{c|}{-6.0} \\ 
\multicolumn{1}{|l|}{\textbf{Qwen3-VL-2b-Instruct}} &
  \multicolumn{1}{c|}{3.23} &
\multicolumn{1}{c|}{-4.56} & 
\multicolumn{1}{c|}{1.85} & 
\multicolumn{1}{c|}{2.21} & 
\multicolumn{1}{c|}{1.74} & 
\multicolumn{1}{c|}{9.33} & 
\multicolumn{1}{c|}{-3.16} & 
\multicolumn{1}{c|}{4.32} & 
\multicolumn{1}{c|}{-8.5} \\ \hline
\multicolumn{10}{|c|}{
\cellcolor[HTML]{E5F7FC}
\textbf{Open source MLLMs (4B - 13B parameters)}} \\ \hline
\multicolumn{1}{|l|}{\textbf{Gemma3-4b}} &
  \multicolumn{1}{c|}{-33.32} &
\multicolumn{1}{c|}{-12.07} & 
\multicolumn{1}{c|}{-4.68} & 
\multicolumn{1}{c|}{-3.94} & 
\multicolumn{1}{c|}{6.09} & 
\multicolumn{1}{c|}{-15.42} & 
\multicolumn{1}{c|}{-2.63} & 
\multicolumn{1}{c|}{2.33} & 
\multicolumn{1}{c|}{-3.0}  \\
\multicolumn{1}{|l|}{\textbf{Chameleon-7b}} &
  \multicolumn{1}{c|}{-15.07} &
\multicolumn{1}{c|}{0.99} & 
\multicolumn{1}{c|}{0.12} & 
\multicolumn{1}{c|}{-2.71} & 
\multicolumn{1}{c|}{-2.98} & 
\multicolumn{1}{c|}{-1.24} & 
\multicolumn{1}{c|}{-8.6} & 
\multicolumn{1}{c|}{-2.65} & 
\multicolumn{1}{c|}{2.0} \\
\multicolumn{1}{|l|}{\textbf{LLaVA-v1.5-7b}} &
  \multicolumn{1}{c|}{-31.95} &
\multicolumn{1}{c|}{-13.3} & 
\multicolumn{1}{c|}{-0.25} & 
\multicolumn{1}{c|}{-8.62} & 
\multicolumn{1}{c|}{-3.73} & 
\multicolumn{1}{c|}{-3.98} & 
\multicolumn{1}{c|}{-1.23} & 
\multicolumn{1}{c|}{0.16} & 
\multicolumn{1}{c|}{-1.0} \\
\multicolumn{1}{|l|}{\textbf{LLaVA-NeXT-Interleave-7b}} &
\multicolumn{1}{c|}{1.28} &
\multicolumn{1}{c|}{-1.23} & 
\multicolumn{1}{c|}{-1.11} & 
\multicolumn{1}{c|}{-3.2} & 
\multicolumn{1}{c|}{0.12} & 
\multicolumn{1}{c|}{-0.38} & 
\multicolumn{1}{c|}{-1.75} & 
\multicolumn{1}{c|}{-0.17} & 
\multicolumn{1}{c|}{9.0}  \\
\multicolumn{1}{|l|}{\textbf{LLaVA-OneVision-7b-SI}} &
\multicolumn{1}{c|}{-39.02} &
\multicolumn{1}{c|}{-11.57} & 
\multicolumn{1}{c|}{-5.55} & 
\multicolumn{1}{c|}{-5.67} & 
\multicolumn{1}{c|}{-3.48} & 
\multicolumn{1}{c|}{-1.12} & 
\multicolumn{1}{c|}{-2.63} & 
\multicolumn{1}{c|}{-1.0} & 
\multicolumn{1}{c|}{-8.0} \\
\multicolumn{1}{|l|}{\textbf{LLaVA-OneVision-7b-OV}} &
  \multicolumn{1}{c|}{-13.64} &
\multicolumn{1}{c|}{-2.96} & 
\multicolumn{1}{c|}{5.17} & 
\multicolumn{1}{c|}{-2.95} & 
\multicolumn{1}{c|}{1.24} & 
\multicolumn{1}{c|}{0.99} & 
\multicolumn{1}{c|}{-2.63} & 
\multicolumn{1}{c|}{-0.5} & 
\multicolumn{1}{c|}{-12.0} \\
\multicolumn{1}{|l|}{\textbf{DeepSeek-VL-7b}} &
  \multicolumn{1}{c|}{-11.06} &
\multicolumn{1}{c|}{1.97} & 
\multicolumn{1}{c|}{1.23} & 
\multicolumn{1}{c|}{-3.2} & 
\multicolumn{1}{c|}{-0.25} & 
\multicolumn{1}{c|}{-0.12} & 
\multicolumn{1}{c|}{-0.35} & 
\multicolumn{1}{c|}{-0.34} & 
\multicolumn{1}{c|}{-10.0}\\
\multicolumn{1}{|l|}{\textbf{Qwen3-VL-8b-Instruct}} &
\multicolumn{1}{c|}{-21.89} &
\multicolumn{1}{c|}{-7.63} & 
\multicolumn{1}{c|}{-7.02} & 
\multicolumn{1}{c|}{-1.97} & 
\multicolumn{1}{c|}{-6.10} & 
\multicolumn{1}{c|}{-4.35} & 
\multicolumn{1}{c|}{-8.77} & 
\multicolumn{1}{c|}{13.95} & 
\multicolumn{1}{c|}{0.00}  \\
\multicolumn{1}{|l|}{\textbf{Monkey-Chat}} &
  \multicolumn{1}{c|}{-3.17} &
\multicolumn{1}{c|}{-1.11} & 
\multicolumn{1}{c|}{-1.10} & 
\multicolumn{1}{c|}{-1.97} & 
\multicolumn{1}{c|}{-1.37} & 
\multicolumn{1}{c|}{-1.24} & 
\multicolumn{1}{c|}{-0.88} & 
\multicolumn{1}{c|}{0.50} & 
\multicolumn{1}{c|}{4.00}  \\ 
\multicolumn{1}{|l|}{\textbf{Idefics2-8b}} &
  \multicolumn{1}{c|}{-5.00} &
\multicolumn{1}{c|}{-5.91} & 
\multicolumn{1}{c|}{2.46} & 
\multicolumn{1}{c|}{0.74} & 
\multicolumn{1}{c|}{-0.74} & 
\multicolumn{1}{c|}{0.62} & 
\multicolumn{1}{c|}{-0.35} & 
\multicolumn{1}{c|}{0.17} & 
\multicolumn{1}{c|}{-2.0} \\
\multicolumn{1}{|l|}{\textbf{InternVL3-8b}} &
  \multicolumn{1}{c|}{-62.16} &
\multicolumn{1}{c|}{-5.55} & 
\multicolumn{1}{c|}{-3.7} & 
\multicolumn{1}{c|}{-0.98} & 
\multicolumn{1}{c|}{-6.84} & 
\multicolumn{1}{c|}{-14.8} & 
\multicolumn{1}{c|}{-13.16} & 
\multicolumn{1}{c|}{-9.13} & 
\multicolumn{1}{c|}{-8.0} \\
\multicolumn{1}{|l|}{\textbf{Idefics-9b-Instruct}} &
  \multicolumn{1}{c|}{-9.19} &
\multicolumn{1}{c|}{-3.45} & 
\multicolumn{1}{c|}{-0.74} & 
\multicolumn{1}{c|}{-0.98} & 
\multicolumn{1}{c|}{3.11} & 
\multicolumn{1}{c|}{-2.98} & 
\multicolumn{1}{c|}{0.18} & 
\multicolumn{1}{c|}{-0.33} & 
\multicolumn{1}{c|}{-4.0}\\
\multicolumn{1}{|l|}{\textbf{Gemma3-12b}} &
  \multicolumn{1}{c|}{-55.78} &
\multicolumn{1}{c|}{-11.33} & 
\multicolumn{1}{c|}{-4.31} & 
\multicolumn{1}{c|}{-5.66} & 
\multicolumn{1}{c|}{-4.48} & 
\multicolumn{1}{c|}{-11.57} & 
\multicolumn{1}{c|}{-5.44} & 
\multicolumn{1}{c|}{-1.99} & 
\multicolumn{1}{c|}{-11.00}\\ \hline
\multicolumn{10}{|c|}{
\cellcolor[HTML]{E5F7FC}
\textbf{Open source MLLMs (\textgreater 13B parameters)}} \\ \hline
\multicolumn{1}{|l|}{\textbf{LLaVA-v1.5-13b}} &
  \multicolumn{1}{c|}{-34.09} &
\multicolumn{1}{c|}{-9.61} & 
\multicolumn{1}{c|}{2.09} & 
\multicolumn{1}{c|}{-7.63} & 
\multicolumn{1}{c|}{-1.24} & 
\multicolumn{1}{c|}{-6.1} & 
\multicolumn{1}{c|}{1.4} & 
\multicolumn{1}{c|}{0.0} & 
\multicolumn{1}{c|}{-13.0} \\
\multicolumn{1}{|l|}
  {\textbf{Qwen3-VL-32b-Instruct}} &
  \multicolumn{1}{c|}{-41.26} &
\multicolumn{1}{c|}{-15.02} & 
\multicolumn{1}{c|}{-6.15} & 
\multicolumn{1}{c|}{-9.36} & 
\multicolumn{1}{c|}{-2.61} & 
\multicolumn{1}{c|}{-6.72} & 
\multicolumn{1}{c|}{-1.23} & 
\multicolumn{1}{c|}{-0.17} & 
\multicolumn{1}{c|}{0.0} \\
\multicolumn{1}{|l|}
  {\textbf{InternVL3-38b}} &
  \multicolumn{1}{c|}{-52.98} &
\multicolumn{1}{c|}{-4.19} & 
\multicolumn{1}{c|}{-0.62} & 
\multicolumn{1}{c|}{-3.94} & 
\multicolumn{1}{c|}{-6.47} & 
\multicolumn{1}{c|}{-20.9} & 
\multicolumn{1}{c|}{-7.54} & 
\multicolumn{1}{c|}{-4.32} & 
\multicolumn{1}{c|}{-5.0} \\
   \hline
\multicolumn{10}{|c|}{
\cellcolor[HTML]{FFFFC7}
\textbf{Proprietary MLLMs}} \\ \hline
\multicolumn{1}{|l|}{\textbf{GPT-5}} &
  \multicolumn{1}{c|}{34.03} &
 \multicolumn{1}{c|}{-4.31} & 
\multicolumn{1}{c|}{7.15} & 
\multicolumn{1}{c|}{-7.14} & 
\multicolumn{1}{c|}{9.83} & 
\multicolumn{1}{c|}{-6.47} & 
\multicolumn{1}{c|}{4.74} & 
\multicolumn{1}{c|}{30.23} & 
\multicolumn{1}{c|}{0.0} \\
\multicolumn{1}{|l|}{\textbf{Gemini 2.5 Pro}} &
  \multicolumn{1}{c|}{-100.76} &
\multicolumn{1}{c|}{-12.19} & 
\multicolumn{1}{c|}{-19.46} & 
\multicolumn{1}{c|}{-10.84} & 
\multicolumn{1}{c|}{-17.53} & 
\multicolumn{1}{c|}{-9.82} & 
\multicolumn{1}{c|}{5.62} & 
\multicolumn{1}{c|}{-35.54} & 
\multicolumn{1}{c|}{-1.0} \\ \hline
\end{tabular}%
}
\end{table*}

\section{Zero-shot vs. CoT output comparison}
\begin{figure*}[htb!]
    \centering
    \includegraphics[width=0.8\textwidth]{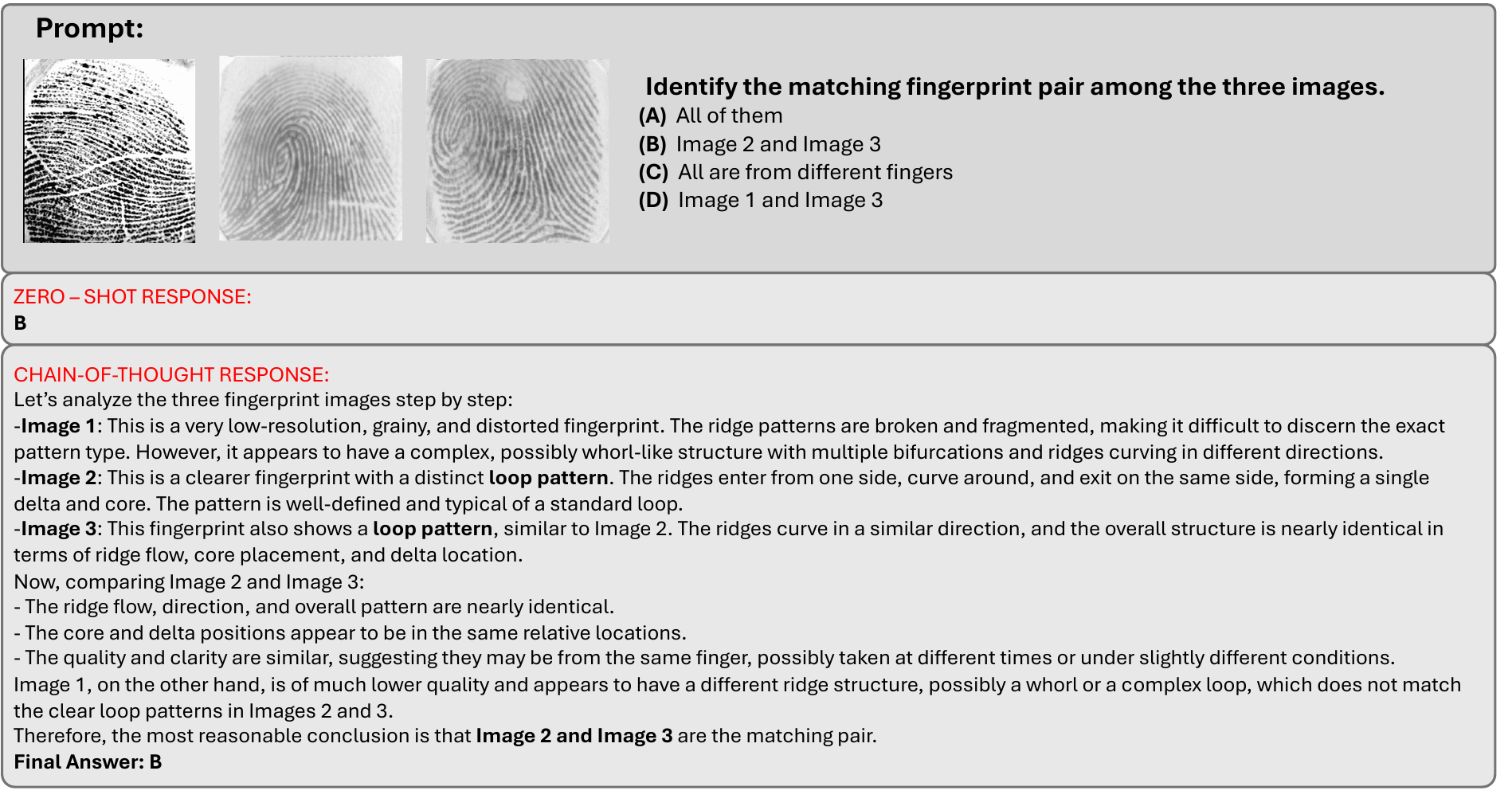}
    \caption{Comparison between responses produced by Qwen3-VL-32b\cite{Qwen3-VL} on the `verification' task from zero-shot and CoT prompting.}
    \label{fig:comparison}
\end{figure*}

Manual review of the CoT response of Qwen3-VL-32b model revealed reasoning inconsistencies (Fig \ref{fig:comparison}). Despite concluding a match, the model assigned incorrect and dissimilar pattern classes to the images, indicating high-level understanding without fine-grained discrimination. This opens the path for domain adaptation that can potentially improve the CoT reasoning capabilities of the models.



\section{Tool Retrieval Questions Generation:}
\label{appendix:a}
Following the approach in \cite{narayan2025facexbench}, we designed a detailed prompt with 10 APIs and a total of 22 functions to generate the \textsc{FPBench} questions. We included APIs covering a wide range of fingerprint understanding tasks, complying with real-world scenarios and systems. We generated a total of 150 questions each from GPT-5 and Gemini 2.5 Pro, out of which 100 were selected. We manually reviewed and rephrased the questions to
maintain diversity and correctness. The detailed prompt for question generation is provided below.


\noindent \textbf{Prompt for Generating Tool Retrieval Questions:}

You are an AI tasked with generating complex, real-world scenario questions to assess a model’s ability to select the correct API and function calls to accomplish nuanced tasks. Use the list of APIs and functions provided below.

\begin{enumerate}
    \item
\textbf{Pattern Classification:} \\
\textbf{api\_name}: \texttt{api\_1} 
\begin{itemize}
    \item \texttt{classify\_pattern} \\
    \textbf{Description}: Predicts the pattern class in a given fingerprint image. \\
    \textbf{Input}: \texttt{np.ndarray} or \texttt{str} - The input fingerprint image. \\
    \textbf{Output}: \texttt{str} - The predicted pattern (`loop', `whorl', or `arch').
    
    \item \texttt{get\_pattern\_probabilities} \\
    \textbf{Description}: Returns probabilities for each pattern class. \\
    \textbf{Input}: \texttt{np.ndarray} - The input fingerprint image. \\
    \textbf{Output}: \texttt{dict} - Probabilities for each pattern class.
    
    \item \texttt{match\_pattern} \\
    \textbf{Description}: Checks whether the fingerprint pair has the same pattern class. \\
    \textbf{Input}: Two \texttt{str} - Two fingerprint pattern classes. \\
    \textbf{Output}: \texttt{bool} - True if fingerprint patterns match, False otherwise.
\end{itemize}

\item \textbf{Matching:} \\
\textbf{api\_name}: \texttt{api\_2}
\begin{itemize}
    \item \texttt{extract\_features} \\
    \textbf{Description}: Extracts fingerprint features from fingerprint images. \\
    \textbf{Input}: \texttt{np.ndarray} or \texttt{str} - The input fingerprint image. \\
    \textbf{Output}: \texttt{np.ndarray} - Feature vector for the fingerprint.
    
    \item \texttt{compare\_features} \\
    \textbf{Description}: Compares two fingerprint feature vectors for a match. \\
    \textbf{Input}: Two \texttt{np.ndarray} - The feature vectors of two fingerprint images. \\
    \textbf{Output}: \texttt{bool} - True if fingerprints match, False otherwise.
    
    \item \texttt{get\_matching\_score} \\
    \textbf{Description}: Get matching score for a given fingerprint image pair. \\
    \textbf{Input}: Two \texttt{str} or \texttt{np.ndarray} - Input fingerprint image pair. \\
    \textbf{Output}: \texttt{int} - Matching score.
    
    \item \texttt{identify\_fingerprint} \\
    \textbf{Description}: Identifies a fingerprint impression by comparing a feature vector to a database. \\
    \textbf{Input}: \texttt{np.ndarray} features and \texttt{dict} database - Feature vector to identify and known features. \\
    \textbf{Output}: \texttt{str} or \texttt{None} - Label of the identified subject/impression/source, or \texttt{None} if no match.
\end{itemize}

\item \textbf{Minutiae:} \\
\textbf{api\_name}: \texttt{api\_3}
\begin{itemize}
    \item \texttt{extract\_minutiae} \\
    \textbf{Description}: Extracts fingerprint minutiae from a given fingerprint image. \\
    \textbf{Input}: \texttt{np.ndarray} or \texttt{str} - The input fingerprint image. \\
    \textbf{Output}: \texttt{np.ndarray} - List of x, y, theta, type of minutiae points.
    
    \item \texttt{plot\_minutiae\_overlay} \\
    \textbf{Description}: Plots minutiae map over fingerprint image. \\
    \textbf{Input}: \texttt{str} or \texttt{np.ndarray} - Input fingerprint image. \\
    \textbf{Output}: \texttt{np.ndarray} - Overlaid minutiae over fingerprint as image.
    
    \item \texttt{get\_minutiae\_count} \\
    \textbf{Description}: Get count of total minutiae, ridge endings, and ridge bifurcations from minutiae points list. \\
    \textbf{Input}: \texttt{np.ndarray} - List of minutiae locations (x, y, theta, type). \\
    \textbf{Output}: Three \texttt{int}s - Number of minutiae, ridge endings, and ridge bifurcations.
\end{itemize}

\item \textbf{Orientation:} \\
\textbf{api\_name}: \texttt{api\_4}
\begin{itemize}
    \item \texttt{get\_orientation\_angles} \\
    \textbf{Description}: Get orientation angles for every \(k \times k\) block in input fingerprint image. \\
    \textbf{Input}: \texttt{str} or \texttt{np.ndarray} and \texttt{int} - Input fingerprint image and block size \(k\). \\
    \textbf{Output}: \texttt{np.ndarray} - 2D array of orientation angles.
    
    \item \texttt{plot\_orientation\_map} \\
    \textbf{Description}: Draw orientation flow map on fingerprint image. \\
    \textbf{Input}: \texttt{str} or \texttt{np.ndarray} and \texttt{np.ndarray} - Input fingerprint image and orientation angles. \\
    \textbf{Output}: \texttt{np.ndarray} - Output image of orientation flow map overlaid on fingerprint.
\end{itemize}

\item \textbf{Sensor Classification:} \\
\textbf{api\_name}: \texttt{api\_5}
\begin{itemize}
    \item \texttt{predict\_sensor\_type} \\
    \textbf{Description}: Predict the type of sensor used to capture the fingerprint image. \\
    \textbf{Input}: \texttt{str} or \texttt{np.ndarray} - Input fingerprint image. \\
    \textbf{Output}: \texttt{str} - Predicted sensor type (`optical', `thermal', `capacitive', or `none').
    
    \item \texttt{get\_sensor\_probabilities} \\
    \textbf{Description}: Returns probability values for each sensor type. \\
    \textbf{Input}: \texttt{str} or \texttt{np.ndarray} - Input fingerprint image. \\
    \textbf{Output}: \texttt{dict} - Probabilities for each sensor type.
\end{itemize}

\item \textbf{Real vs Synthetic Classification:} \\
\textbf{api\_name}: \texttt{api\_6}
\begin{itemize}
    \item \texttt{predict\_real\_synthetic} \\
    \textbf{Description}: Predict if the input fingerprint image is real or synthetic. \\
    \textbf{Input}: \texttt{str} or \texttt{np.ndarray} - Input fingerprint image. \\
    \textbf{Output}: \texttt{str} - Predicted class (`real' or `synthetic').
    
    \item \texttt{get\_real\_probabilities} \\
    \textbf{Description}: Returns probabilities for each class. \\
    \textbf{Input}: \texttt{str} or \texttt{np.ndarray} - Input fingerprint image. \\
    \textbf{Output}: \texttt{dict} - Probabilities for `real' and `synthetic'.
    
    \item \texttt{detect\_bonafide} \\
    \textbf{Description}: Detect if the given fingerprint is bonafide and not spoof. \\
    \textbf{Input}: \texttt{np.ndarray} - Input fingerprint image. \\
    \textbf{Output}: \texttt{bool} - True if bonafide, False otherwise.
\end{itemize}

\item \textbf{ACE-V Analysis:} \\
\textbf{api\_name}: \texttt{api\_7}
\begin{itemize}
    \item \texttt{prepare\_ace\_sheet} \\
    \textbf{Description}: Prepare ACE-V style sheet with desired fields from a pair of fingerprint images. \\
    \textbf{Input}: Input fingerprint image. \\
    \textbf{Output}: \texttt{dict} - Output ACE-V sheet.
    
    \item \texttt{compare\_fingerprint\_ace} \\
    \textbf{Description}: Get `individualization' or `exclusion' decision from the given ACE-V sheet. \\
    \textbf{Input}: \texttt{dict} - ACE sheet. \\
    \textbf{Output}: \texttt{str} - Decision for comparison (`individualization' or `exclusion').
\end{itemize}

\item \textbf{Fingerprint Enhancement:} \\
\textbf{api\_name}: \texttt{api\_8}
\begin{itemize}
    \item \texttt{enhance\_image} \\
    \textbf{Description}: Enhance given fingerprint image for feature extraction. \\
    \textbf{Input}: \texttt{str} or \texttt{np.ndarray} - Input fingerprint image. \\
    \textbf{Output}: \texttt{np.ndarray} - Enhanced fingerprint image.
\end{itemize}

\item \textbf{Fingerprint Segmentation:} \\
\textbf{api\_name}: \texttt{api\_9}
\begin{itemize}
    \item \texttt{segment\_palm\_print} \\
    \textbf{Description}: Segment the given palm print image into five fingerprint segments. \\
    \textbf{Input}: \texttt{str} or \texttt{np.ndarray} - Input palm print image. \\
    \textbf{Output}: Five \texttt{np.ndarray} - Five fingerprint images, one for each finger. Returns \texttt{None} if no fingerprint detected.
\end{itemize}

\item \textbf{Fingerprint Quality:} \\
\textbf{api\_name}: \texttt{api\_10}
\begin{itemize}
    \item \texttt{get\_quality\_score} \\
    \textbf{Description}: Get quality score for the given fingerprint image. \\
    \textbf{Input}: \texttt{str} or \texttt{np.ndarray} - Input fingerprint image. \\
    \textbf{Output}: \texttt{int} - Integer quality score from 1 to 100.
\end{itemize}
\end{enumerate}

\noindent \textbf{Guidelines for Generating Questions:}

\begin{itemize}
    \item \textbf{Scenario Realism:} Design questions reflecting realistic application scenarios where multiple APIs must be used in sequence or combined to achieve the correct outcome. Each question should require 3–5 function calls.
    \item \textbf{Functional Complexity:} Ensure each question involves varied functions across multiple APIs without relying on the same set of functions every time.
    \item \textbf{Logical Flow:} Each question should suggest a sequence that logically flows with the task requirements. Clarify steps needed for functions that build upon each other to reach the final answer.
\end{itemize}

\noindent \textbf{Guidelines for Generating Options:}
\begin{itemize}
    \item \textbf{Complete API Chains:} Provide four option chains, each specifying a complete sequence of API function calls in the correct order. One sequence should be correct; the others should be logically incorrect but plausible.
    \item \textbf{Logical Plausibility of Distractors:} Distractors should appear reasonable and require reasoning to eliminate.
    \item \textbf{Randomized Answer Positioning:} Shuffle options so the correct answer appears randomly in position A, B, C, or D.
\end{itemize}

\noindent \textbf{Example Question:} In an airport security system, a fingerprint is enhanced and checked for bonafide print and, if yes, verified against the stored database. Which API sequence should be applied? 

\begin{enumerate}[A.]
    \item \texttt{api\_8-enhance\_image, api\_2-extract\_features, api\_6-detect\_bonafide, api\_2-identify\_fingerprint}
    \item \texttt{api\_8-enhance\_image, api\_2-extract\_features, api\_2-identify\_fingerprint, api\_6-detect\_bonafide}
    \item \texttt{api\_8-enhance\_image, api\_6-detect\_bonafide, api\_2-extract\_features, api\_2-identify\_fingerprint}
    \item \texttt{api\_6-detect\_bonafide, api\_2-extract\_features, api\_2-identify\_fingerprint, api\_8-enhance\_image}
\end{enumerate}

\noindent Correct Answer: 
\texttt{C. api\_8-enhance\_image, api\_6-detect\_bonafide, api\_2-extract\_features, api\_2-identify\_fingerprint}
 

\end{document}